\documentclass{article}

\usepackage[nonatbib, final]{neurips_2024} 
\usepackage{xr}
\usepackage{hyperref}

\usepackage[utf8]{inputenc} 
\usepackage[T1]{fontenc}    
\usepackage{microtype}      

\usepackage{amsmath}    
\usepackage{amsfonts}   
\usepackage{nicefrac}   

\usepackage{array}        
\usepackage{booktabs}     
\usepackage{multirow}     
\usepackage{graphicx}     
\usepackage{float}        
\usepackage{pdflscape}    
\usepackage{subcaption}   

\usepackage[table]{xcolor} 
\usepackage{tkz-euclide}
\usepackage{tikz-3dplot}
\usetikzlibrary{calc}

\usepackage{url}        
\usepackage{setspace}   
\renewcommand{\arraystretch}{1.1} 

\usepackage[ruled,vlined,algo2e]{algorithm2e} 

\usepackage[
url=false,
  doi=true,
  isbn=false,
  sorting=none,
  eprint=false]{biblatex}

\AtEveryBibitem{%
  \clearfield{url}%
  \clearfield{urldate}%
  \clearfield{isbn}%
  \clearfield{issn}%
  \clearfield{eprint}%
  \clearfield{month}
  \clearfield{eprintclass}%
  \clearfield{eprinttype}%
  \clearlist{language}%
  \clearfield{note}
}
  
\usepackage[table]{xcolor} 
\usepackage{tkz-euclide}
\usepackage{tikz-3dplot}
\usetikzlibrary{calc}

\usepackage{url}        
\usepackage{setspace}   
\renewcommand{\arraystretch}{1.1} 

\providecommand{\degree}{^\circ}

\usepackage[ruled,vlined,algo2e]{algorithm2e}

\definecolor{high_cov}{HTML}{e8ff99}  
\definecolor{low_cov}{HTML}{f6c193}   

\definecolor{low}{HTML}{76f013}  
\definecolor{high}{HTML}{006400}  

\newcommand*{\opacity}{90}

\newcommand{\gradientAMR}[1]{
    \pgfmathparse{#1<0.12 ? 0.12 : (#1>0.26 ? 0.26 : #1)}
    \xdef\clampedval{\pgfmathresult}
    \pgfmathparse{int(round(100*(1 - (\clampedval-0.12)/(0.26-0.12))))}
    \xdef\tempa{\pgfmathresult}
    \cellcolor{high!\tempa!low!\opacity}\color{white}#1
}

\newcommand{\gradientCov}[1]{
    \pgfmathparse{#1<20 ? 20 : (#1>80 ? 80 : #1)}
    \xdef\clampedval{\pgfmathresult}
    \pgfmathparse{int(round(100*(\clampedval-20)/(80-20)))}
    \xdef\tempa{\pgfmathresult}
    \cellcolor{high_cov!\tempa!low_cov!\opacity}#1
}

\definecolor{my_color}{HTML}{D14D3D}

\newcounter{docpart}
\def\savestatus{%
  \newwrite\tempfile%
  \immediate\openout\tempfile=docstatus\arabic{docpart}.dat%
  \immediate\write\tempfile{\thesection}%
  \immediate\write\tempfile{\theequation}%
  \immediate\closeout\tempfile%
}
\newcounter{olddocpart}

\DontPrintSemicolon
\SetKwInput{Input}{Input}
\SetKwInput{Output}{Output}
\SetKw{KwBy}{by}
\SetKw{To}{ to }
\SetKw{print}{print }
\title{Generating Cyclic Conformers with Flow Matching \\ in Cremer-Pople Coordinates}

\newcommand{\sym}[1]{\textsuperscript{#1}}

\author{%
  Luca Schaufelberger\sym{*†‡} \\
  \texttt{schaluca@ethz.ch}
  \And
  Aline Hartgers\sym{*†‡} \\
  \texttt{ahartgers@ethz.ch}
  \And
  Kjell Jorner\sym{†‡}\\
  \texttt{kjorner@ethz.ch}
}

\newcommand{\name}{{PuckerFlow}}

\begin{document}
\begin{refsection}

\maketitle

\begingroup
\renewcommand{\thefootnote}{\fnsymbol{footnote}}
\footnotetext[1]{Equal contribution.}
\footnotetext[2]{Institute of Chemical and Bioengineering, Department of Chemistry and Applied Biosciences, ETH Zurich.}
\footnotetext[3]{NCCR Catalysis, Switzerland.}
\endgroup

\begin{abstract}

   Cyclic molecules are ubiquitous across applications in chemistry and biology. Their restricted conformational flexibility provides structural pre-organization that is key to their function in drug discovery and catalysis.   However, reliably sampling the conformer ensembles of ring systems remains challenging.
Here, we introduce \name{}, a generative machine learning model that performs flow matching on the Cremer-Pople space, a low-dimensional internal coordinate system capturing the relevant degrees of freedom of rings. 
Our approach enables generation of valid closed rings by design
and demonstrates strong performance in generating conformers that are both diverse and precise. 
We show that \name{} outperforms other conformer generation methods on nearly all quantitative metrics and illustrate the potential of \name{} for ring systems relevant to chemical applications, particularly in catalysis and drug discovery.
This work enables efficient and reliable conformer generation of cyclic structures, paving the way towards modeling structure-property relationships and the property-guided generation of rings across a wide range of applications in chemistry and biology.

\end{abstract}

\section{Introduction}

\begin{figure}[t!]
    \centering
    \includegraphics[width=\textwidth]{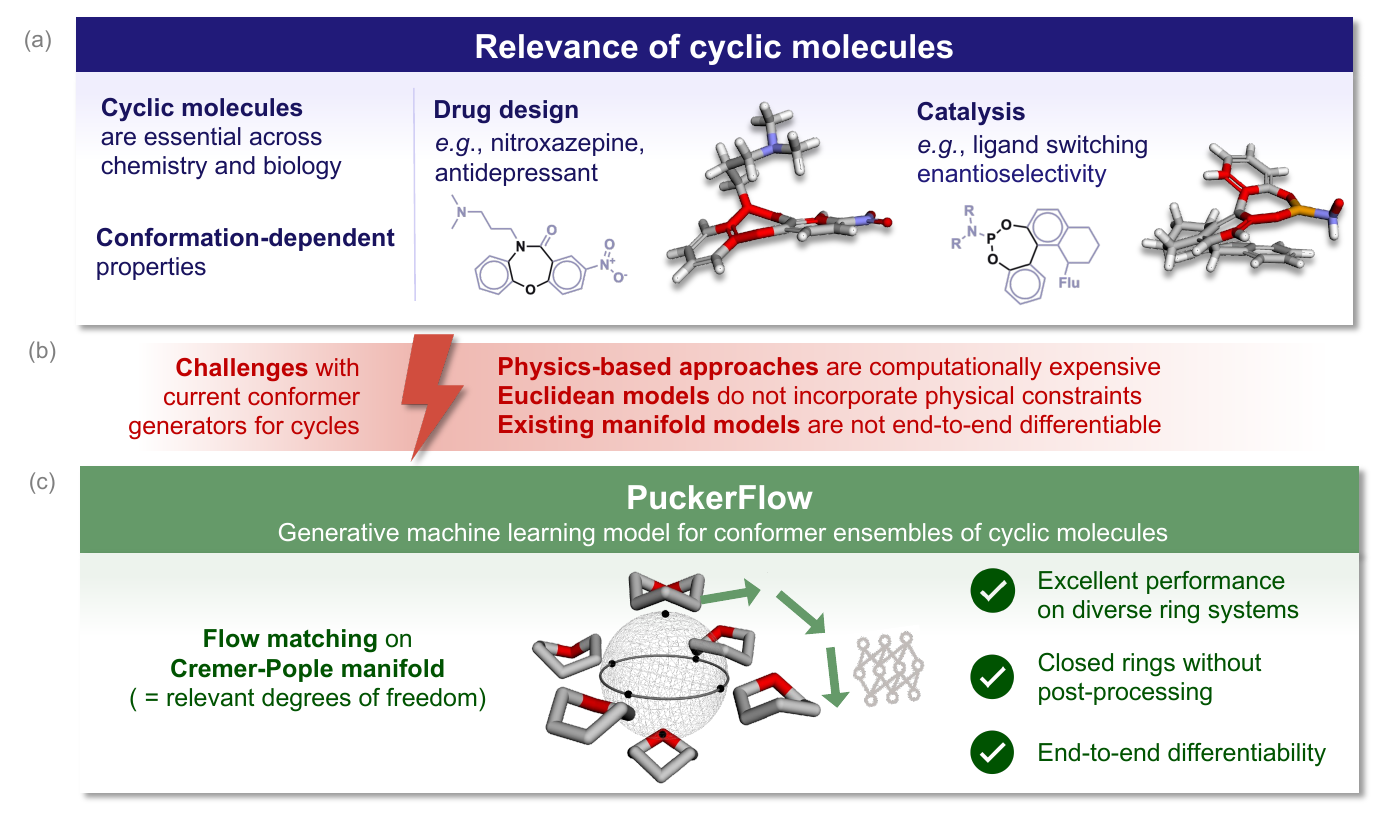}
    \caption{Overview of the presented work. (a) Cyclic molecules are omnipresent in applications such as drug design and catalysis \cite{pizzolato_phosphoramidite-based_2020}, with their conformational space crucial to their function (Flu = fluorene). (b) Current approaches show limited performance on cyclic structures, or can produce non-closed rings. (c) Here, we introduce \name{}, a generative machine learning model that operates only on the relevant degrees of freedom of ring systems and shows state-of-the-art performance on diverse ring systems. 
}
    \label{fig:overview}
\end{figure}

Cyclic motifs are fundamental to molecular function. For example, over 99\% of bioactive molecules feature a ring system as a core structural element \cite{ertl_magic_2022}. Cyclic structures possess unique structural characteristics by offering a compromise between pre-organization and flexibility, which orients functional groups and lowers the entropic cost when interacting with molecules and proteins \cite{hayward_strategies_2024, horton_exploring_2002}.
The unique structural characteristics of cyclic systems make them the central building blocks for a wide range of metabolites (\textit{e.g.}, ATP and saccharides) \cite{walsh_eight_2018}. Their functional properties are furthermore crucial in the fields of drug discovery and catalysis (see Fig. \ref{fig:overview}a).
In drug discovery, ring systems have shown particular promise in targeting low-druggability proteins, especially those with shallow or extended binding surfaces such as protein-protein interaction interfaces \cite{driggers_exploration_2008, ji_cyclic_2024, villar_how_2014}. In catalysis, many catalysts incorporate cyclic substructures that stabilize the transition state, thereby modulating reaction rates and selectivity \cite{molecules28052234, Gurka2022}.

Small- and medium-sized rings are of special interest, as their conformational flexibility is strongly reduced by internal strain. These energetic constraints provide structural pre-organization to such rings, which can \textit{e.g.}, enhance their ability to bind selectively to biological targets. The nature of the strain varies with ring size: torsional and angular strain dominate in smaller rings, whereas transannular interactions, \textit{i.e.}, across the ring, become more relevant in medium-sized rings \cite{anslyn_modern_2006}. Medium-sized rings have, for example, been shown to induce improved binding affinities \cite{romines_use_1995}, oral bioavailability \cite{veber_molecular_2002} and cellular permeability \cite{kwon_quantitative_2007} compared to macrocycles \cite{reyes_construction_2021}.
For small- and medium-sized rings, the conformer space is characterized by their \textit{puckering}, \textit{i.e.}, the out-of-plane displacement of the ring atoms, with specific puckering modes playing crucial roles in tuning the molecular properties \cite{chan_understanding_2021}.

Reliably and efficiently characterizing the conformer landscape of ring systems is essential to understanding their function. However, the strong coupling of the torsional angles in ring systems makes this a particularly challenging task. 
Several computational methods have been developed that can be used to sample conformers of ring systems. 
Ideally, a conformer generator should both sample the entire conformer space (high recall), while producing structures that are close to energetic minima (high precision).
Molecular dynamics (MD) and enhanced sampling methods such as metadynamics \cite{laio_escaping_2002, pracht_automated_2020} provide physics-based exploration of the conformer space, but are computationally expensive.
Rule-based approaches \cite{rihon_puckers_2025, hawkins_conformer_2010}, like RDKit’s ETKDG \cite{riniker_better_2015, wang_improving_2020} offer fast conformer generation but show limited precision, \textit{i.e.}, closeness to the reference conformers in terms of RMSD.
Recently, generative machine learning models based on diffusion and flow matching have emerged as a highly promising avenue for conformer generation \cite{jing_torsional_2022, wang_swallowing_2024}, 
surpassing cheminformatics tools in precision and outperforming molecular dynamics in computational efficiency. Diffusion models generate samples by reversing a gradual noising process, whereas flow matching uses a learned vector field that deterministically transports noise toward the data distribution.

Euclidean generative models \cite{xu_geodiff_2021, wang_swallowing_2024}, which operate directly on atomic positions, often require large training datasets (\textasciitilde $10^7$--$10^8$ conformers \cite{axelrod_geom_2022}) and cannot incorporate chemical constraints \cite{yim_diffusion_2024} such as ring closure (see Fig. \ref{fig:overview}b). 
%
On the other hand, manifold generative models \cite{yim_diffusion_2024, jing_torsional_2022} that formulate the generation process on internal coordinates instead of Cartesian coordinates, reduce the number of degrees of freedom and thereby offer potentially higher data efficiency and the straightforward incorporation of chemically meaningful constraints. 
One such manifold approach is the torsional diffusion model by Jing \textit{et al.} \cite{jing_torsional_2022}, which generates conformers by only diffusing over the exocyclic torsional degrees of freedom (dihedral angles), while keeping local structures such as ring systems rigid. 

Despite the promising results of the current manifold diffusion models, they are inherently limited in their application as they do not allow sampling of the complex conformer ensembles of rings, limiting their usability \cite{ganea_geomol_2021}. For example, torsional diffusion relies on RDKit's ETKDG algorithm to sample cyclic substructures \cite{jing_torsional_2022}, and the recently introduced RINGER model for cyclic peptides performs diffusion on the endocyclic torsion angles, leading to non-closed rings that need postprocessing to enforce chemical validity \cite{grambow_accurate_2024}. This reliance on post-hoc correction breaks end-to-end differentiability, with consequences such as preventing the use of the model in gradient-based reward optimization for downstream applications \cite{domingo_i_enrich_adjoint_2025}.

Here, we introduce \name{}, an equivariant generative machine learning model that performs flow matching directly on the most relevant internal degrees of freedom of cyclic structures.
By restricting the generation process to the Cremer-Pople internal coordinate system \cite{cremer_general_1975}, which captures the puckering of rings in a low-dimensional space, our model produces conformers with closed rings without postprocessing and shows strong performance in generating diverse and precise conformers (see Fig. \ref{fig:overview}c).
A key element of \name{} is a novel manifold-adapted cyclic Fourier filter that enables the model to handle rings of varying sizes within a unified framework.
\name{} demonstrates state-of-the-art performance on a diverse set of ring structures \cite{folmsbee_systematic_2023}, and is able to fully sample their conformer space. 
Our method outperforms both Euclidean generative models and RDKit's ETKDG across a broad range of metrics, demonstrating its capacity as a reliable generator of cyclic conformers. 
We highlight the capabilities of \name{} in sampling rings relevant for chemical and pharmaceutical applications across different ring sizes and diverse heteroatom compositions. Being computationally efficient, \name{} is suited for high-throughput discovery campaigns of cyclic scaffolds, promoting sustainable use of compute resources in drug discovery and catalysis.
Our method is fully compatible with complementary manifold generative models, such as torsional diffusion \cite{jing_torsional_2022} for sampling the exocyclic torsion angles and DiffDock \cite{corso_diffdock_2022} for binding pose prediction, allowing  future integration of \name{} into molecular protein docking. 
Overall, \name{} paves the way towards reliable sampling of cyclic conformers for a wide range of applications across chemistry and biology.

\section{Results and Discussion}
\subsection{Modeling in Cremer-Pople coordinates captures the essential degrees of freedom}
\begin{figure}[tbp]
    \centering
    \includegraphics[width=1.0\textwidth]{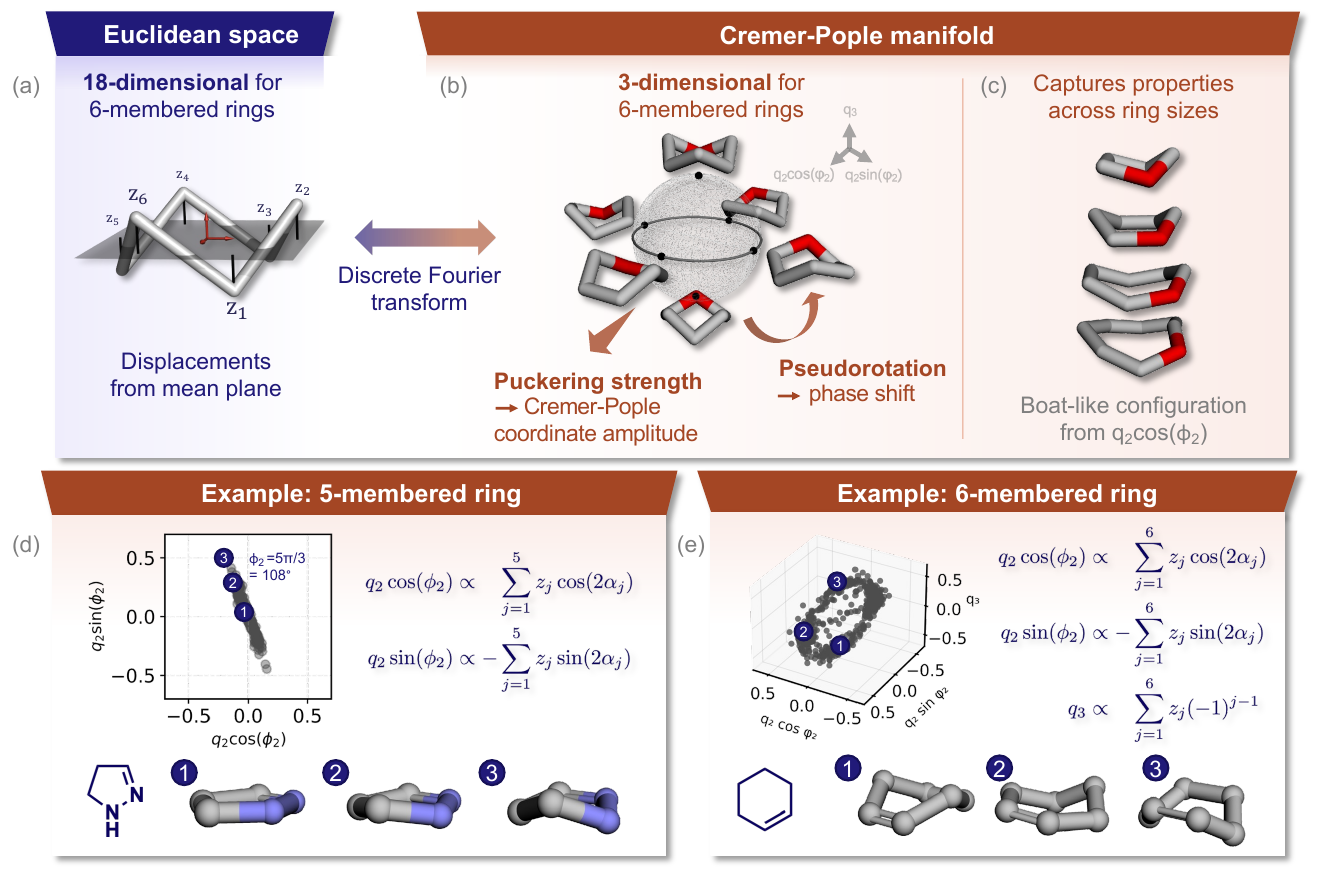}
    \caption{Depiction of the Cremer-Pople coordinates that compactly describe the relevant degrees of freedom of cyclic systems. (a) The puckering of a ring can be quantified by the displacements $z$ from the mean plane (exaggerated for visual clarity), with the coordinate system shown in red. (b) The Cremer-Pople coordinates are a low-dimensional representation that capture the essential degrees of freedom. (c) A single non-zero Cremer-Pople coordinate $q_2 cos(\phi_2)$ describes conformer properties across different ring size. (d-e) Example Cremer-Pople coordinate distributions for (d) a five-membered ring (2-pyrazoline) and (e) a six-membered ring (cyclohexene), with ring geometries extracted from several conformers. We show selected Euclidean structures that illustrate the characteristic conformer distributions in Cremer-Pople space.}
    \label{fig:methods_CP_explanation}
\end{figure}

\name{} efficiently samples conformers of ring systems by performing flow matching (\textit{vide infra}) on a special set of internal coordinates that compactly captures the essential degrees of freedom of ring systems: the Cremer-Pople coordinates \cite{cremer_general_1975}. The Cremer-Pople coordinates characterize the out-of-plane displacements $z$ of the $N$ atoms of a ring, often referred to as \textit{puckering}  (see Fig. \ref{fig:methods_CP_explanation}a). The Cremer-Pople coordinates constitute a manifold, \textit{i.e.}, a lower-dimensional structure embedded in a higher-dimensional space. In this way, the $3N$ Euclidean degrees of freedom in the Cartesian space are reduced to the most relevant $N-3$ degrees of freedom corresponding to the Cremer-Pople coordinates. For example, the coordinate space of a six-membered ring is 18-dimensional in Cartesian coordinates, but just three-dimensional in Cremer-Pople space.

The Cremer-Pople coordinates are constructed as a discrete Fourier transform of the atomic displacements $z$ from the mean plane of the ring \cite{cremer_general_1975}, as visualized in Fig. \ref{fig:methods_CP_explanation}a.  Each pair of Cremer-Pople coordinates ($q_m\cos\phi_m $, $q_m\sin\phi_m)$ of order $m$ captures the amplitude $q_m$ and phase $\phi_m$ of a sinusoidal wave with $m$ maxima. 

Formally, they are defined as \cite{cremer_general_1975}:
\begin{subequations}\label{eq:cp_formula}
\begin{align}
q_m \cos(\phi_m) =  &\ \ \ \  (2 / N)^{1 / 2} \sum_{j=1}^N z_j \cos (m\alpha_j) \\
q_m \sin(\phi_m)  =  & -(2 / N)^{1 / 2} \sum_{j=1}^N z_j \sin (m\alpha_j). \\
&\text{with } m=2,3, \ldots,\lfloor(N-1) / 2\rfloor \nonumber \\
&\text{and } \alpha_j= 2 \pi (j-1) / N\nonumber
\end{align}
\end{subequations}

For even-membered rings, there is an additional Cremer-Pople coordinate:

\begin{equation}\label{eq:cp_formula_odd}
   q_{N / 2} = \ (1/N)^{1 / 2} \sum_{j=1}^N(-1)^{j-1} z_j. 
\end{equation}

Here, the phase $\phi_{N/2}$ is redundant because its effect is equivalent to modifying the amplitude $q_{N/2}$ \cite{cremer_general_1975}.  
The total puckering amplitude of a ring is characterized by $Q = \sqrt{\sum_m {q_m}^2}$. For planar rings, all puckering amplitudes $q_m$ are zero.
Further details on the construction of Cremer-Pople coordinates are given in SI \ref{SI:cart_to_cp}.

Like other internal coordinate representations \cite{oenen_determining_2024,sittel_principal_2014,vaidehi_internal_2015}, the Cremer-Pople coordinates are designed to describe the most relevant degrees of freedom \cite{paoloni_interplay_2021} and capture chemically meaningful structural transformations.
For example, pseudorotations \cite{kilpatrick_thermodynamics_1947,paoloni_interplay_2021} of cyclic structures, \textit{i.e.}, concerted, low-energy interconversion between puckered conformers in rings, correspond to a shift in the phase $\phi_m$ with constant puckering amplitudes $q_m$ \cite{paoloni_potential-energy_2019} (see Fig. \ref{fig:methods_CP_explanation}b). 
The Cremer-Pople coordinates also capture essential conformer characteristics that are transferable across ring sizes, with, \textit{e.g.}, the first puckering coordinate $q_2 \cos(\phi_m)$ describing generalized boat-like conformers as visualized in Fig. \ref{fig:methods_CP_explanation}c.

Figs. \ref{fig:methods_CP_explanation}d-e illustrate the relationship between ring conformer distributions in Cremer-Pople coordinates and molecular conformers in Cartesian space for five- and six-membered  rings, for which the Cremer-Pople space is two- and three-dimensional, respectively. 
These distributions were extracted from molecular conformer ensembles and correspond to instances of the rings as substructures of different molecules.
The five-membered 2-pyrazoline ring (Fig. \ref{fig:methods_CP_explanation}d) exhibits one main puckered mode, \textit{i.e.}, clusters in the conformer distribution in Cremer-Pople space. The mode exhibits a varying amplitude at a fixed phase angle $\phi_2 = 5\pi/3 +k\cdot\pi = 108 ^\circ + k\cdot180^\circ$ with $k\in\mathbb{Z}$, which corresponds to an envelope conformer where four atoms lie approximately in one plane while the fifth atom (the \textit{flap}) is displaced from this plane. The magnitude of this displacement depends on the amplitude $q_2$, with \textit{e.g.}, conformer \textit{1} being planar and conformer \textit{3} showing the strongest puckering (see Fig. \ref{fig:methods_CP_explanation}d).
The conformer distribution of the six-membered cyclohexene ring (Fig. \ref{fig:methods_CP_explanation}e) forms a tilted two-dimensional plane embedded in the three-dimensional Cremer-Pople space, where transformations between the conformers correspond to pseudorotations. 
The most stable conformer of cyclohexene is the half-chair (conformer \textit{2}), while a second mode corresponds to the boat form (conformers \textit{1} and \textit{3}). \cite{jensen_conformational_1969}.

These examples highlight that the Cremer-Pople coordinates compactly describe the relevant conformer characteristics in a low-dimensional subspace, making them highly suitable for generative modeling.

\begin{figure}[t!]
    \centering
    \includegraphics[width=\textwidth]{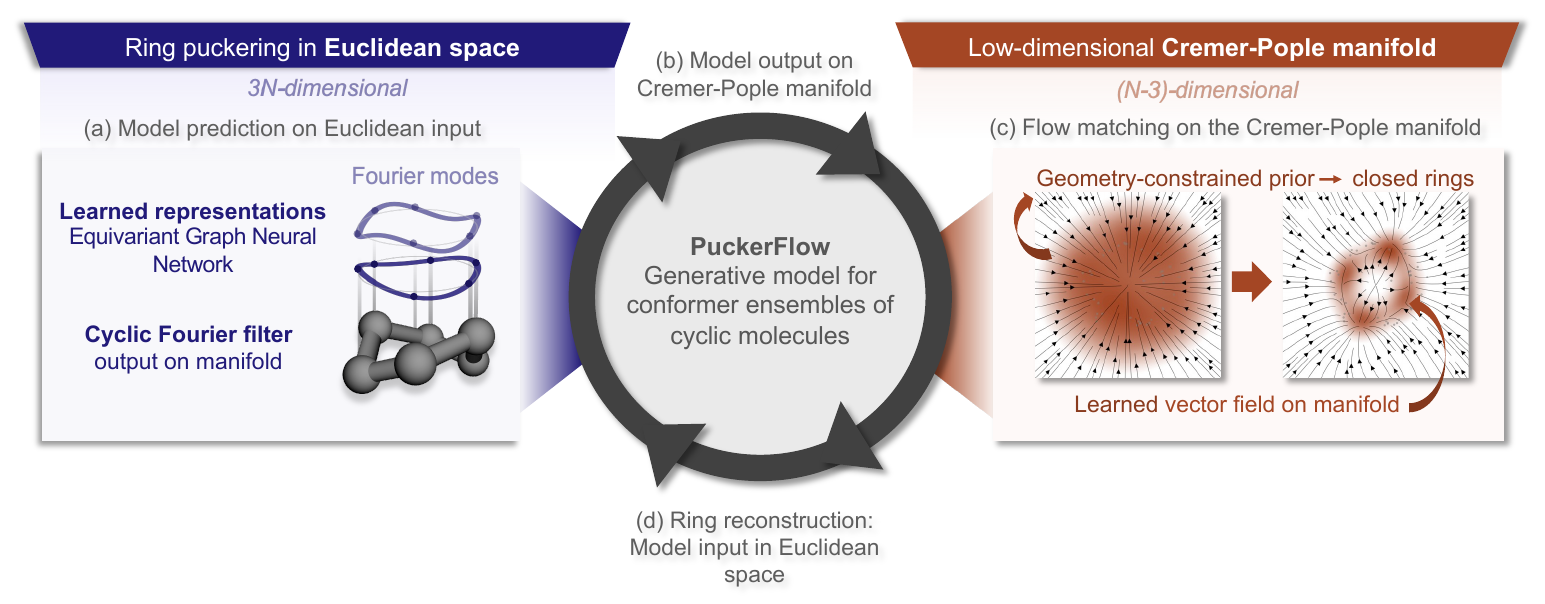}
    \caption{Overview of \name{}, which  performs flow matching on the low-dimensional Cremer-Pople space. (a) Learned atomic representations are passed to the \textit{cyclic Fourier filter} output layer introduced in this work that  (b) produces the symmetry-adapted output in Cremer-Pople coordinates. To train \name{}, (c) we introduce a geometry-informed prior that ensures valid closed rings, and predict flow updates on the low-dimensional Cremer-Pople space, (d) which is used to reconstruct the three-dimensional geometry.}
    \label{fig:methods_model_overview}
\end{figure}

\subsection{Flow matching enables geometry-aware ring generation}\label{sec:flow_matching_method}
\name{} is based on the flow matching paradigm \cite{lipman_flow_2022}, where the model learns a time-dependent vector field that transforms a simple prior distribution (at time $t=0$) into the target distribution (at $t=1$) according to an ordinary differential equation (ODE, see SI~\ref{SI:generative_modeling} for details). Instead of in Euclidean space as conventionally done, flow matching is performed on the Cremer-Pople space in this work. 
The model generates conformers starting from a random sample from the prior distribution in Cremer-Pople space ($\mathbf{x}_{t=0}$). This sample is then iteratively refined into a ring conformer ($\mathbf{x}_{t=1}$) over a certain number of time steps. The first step of this iterative cycle is to (1) convert the values from Cremer-Pople space ($\mathbf{x}_t$) into the corresponding coordinates in Euclidean space. These Euclidean coordinates serve as input to an equivariant 3D graph neural network (see Fig. \ref{fig:methods_model_overview}a) that (2) outputs its current prediction $\mathbf{\tilde{x}}_1$ of the final point in Cremer-Pople space (see Fig. \ref{fig:methods_model_overview}b) using a specially design cyclic \textit{Fourier} output layer (\textit{vide infra}). This prediction $\mathbf{\tilde{x}}_1$ is (3) used together with the current set of coordinates $\mathbf{x}_t$, to produce an updated point in Cremer-Pople space ($\mathbf{x}_{t+\Delta t}$). 
This corresponds to a flow in a (time-dependent) vector field (see Fig. \ref{fig:methods_model_overview}c), where the point $\mathbf{x}_t$ is carried a certain distance in the direction of $\mathbf{\tilde{x}}_1$ according to the corresponding ODE. The new point in Cremer-Pople space is then again back-converted into Euclidean space and the cycle starts anew (back to step 1). These incremental updates lead to the final output structure $\mathbf{x}_1$ at $t=1$ . The detailed procedure for both generation and training are provided as pseudocode in SI~\ref{SI:pseudocode}.

The conversion from Cremer-Pople to Euclidean space (Fig. \ref{fig:methods_model_overview}d) follows the approach of Cremer \cite{cremer_calculation_1990}. It uses precomputed bond lengths and angles from the training data, parameterized by bond type, element, and ring size. This approximation proves sufficient for accurate geometric reconstruction (see SI~\ref{SI:data_preprocessing}) and is detailed in SI~\ref{SI:cp_to_cart}. Since very large Cremer-Pople amplitudes yield unphysical bond lengths (see SI~\ref{SI:cp_feasibility}), ensuring valid rings during training requires careful modeling design choices. This motivates our choice of flow matching \cite{lipman_flow_2022} over diffusion models \cite{ho_denoising_2020}: while the Gaussian prior of diffusion models samples potentially invalid regions, flow matching allows for the design of a geometry-informed prior. In \name{}, we define an amplitude-bounded prior (see SI~\ref{SI:prior_distribution}) that, combined with flow matching's straight conditional training paths and the convexity of valid Cremer-Pople space, ensures valid rings throughout training.

A key challenge is to design a model whose output preserves the inherent symmetries of the Cremer-Pople space and which adapts to its varying dimensionality across ring sizes (see SI~\ref{SI:model_architecture} and~\ref{SI:additional_results} for the full architecture and parameters).
Specifically, for $N$ ring atoms, the model output needs to be $(N-3)$-dimensional, with each Cremer-Pople coordinate being a pseudoscalar, \textit{i.e.}, invariant under rotation but changing sign under parity inversion. To address this, we introduce the \textit{cyclic Fourier filter} as the output layer (see Fig. \ref{fig:methods_model_overview}a), which processes the atom embeddings learned from an equivariant neural network \cite{geiger_e3nn_2022, jing_torsional_2022} to produce the model output. 
The cyclic Fourier filter layer places $N$ equivariant tensor filters on the mean plane ($z=0$), which are each convolved with the learned representations of all other ring atoms  (see SI~\ref{SI:cyclic_fourier_filter} for details).
The outputs of this convolution are then passed through a discrete Fourier transform to produce the flow vector in the Cremer-Pople space, leading to the required $(N-3)$-dimensional output. Hence, our model output is dynamically scaled to the required Cremer-Pople dimensionality for any given input ring size.

\subsection{Data and benchmarking methodology}\label{sec:data_methodology}
To evaluate the performance of \name{}, we build on the ring puckering data of Folmsbee \textit{et al.} \cite{folmsbee_systematic_2023}, which contains molecules extracted from a variety of experimental and computational data sources (Crystallography Open Database \cite{vaitkus_workflow_2023, merkys_graph_2023, vaitkus_validation_2021, quiros_using_2018, merkys_it_2016, grazulis_computing_2015, grazulis_crystallography_2012, grazulis_crystallography_2009, downs_american_2003}, Pitt Quantum Repository \cite{re3dataorg_pitt_2020}, ZINC \cite{irwin_zinc_2005},  and the Platinum ligand database \cite{pires_platinum_2015}, comprising approximately 252,000 organic compounds, with their lowest-energy conformers obtained using the CREST conformer sampling program \cite{pracht_automated_2020}. 
In this work, we focus on small- and medium-sized rings (five- to eight-membered rings), which are highly prevalent in chemistry \cite{reyes_construction_2021} and for which Cremer-Pople coordinates provide an effective geometric description. 
These ring sizes dominate the dataset, comprising more than 99\% of all rings. 
We exclude rings containing radical electrons as well as aromatic rings, since their (limited) distortions from planarity are predominantly induced by ring fusion or exocyclic atoms rather than intrinsic puckering (see details on preprocessing in SI~\ref{SI:data_preprocessing}). This process yields a dataset of 15,204 conformers, coming from 419 unique rings. As we focus on the intrinsic puckering of rings in this work, we train all models only on the cyclic systems without substituents. As shown in Sec. \ref{sec:exocyclic_subst}, \name{} can be extended to also generate exocyclic substituents by integrating it with RDKit's ETKDG or torsional diffusion.
While in this work, we target monocyclic ring systems, our approach can also be applied to fused and spiro rings, where the Cremer-Pople coordinates can be determined for each component ring separately \cite{flachsenberg_ringdecomposerlib_2017}.

We benchmark \name{} against RDKit's (ET)KDG method \cite{riniker_better_2015}, a widely used cheminformatics algorithm, and two generative models, GeoDiff \cite{xu_geodiff_2021} and MCF \cite{wang_swallowing_2024}. MCF currently achieves state-of-the-art performance for conformer sampling on GEOM-QM9 and GEOM-Drugs \cite{axelrod_geom_2022}. While GeoDiff is equivariant by design, MCF does not enforce rotational equivariance, instead learning it implicitly from the data. For RDKit, we benchmark the standard ETKDG implementation, the KDG version, and an extended ETKDG variant that introduces a torsional-angle potential for small rings (\textit{Small Cycles}) \cite{wang_improving_2020}. The \textit{ET} in ETKDG stands for Experimental Torsion, reflecting torsion-angle preferences derived from small-molecule crystal structures in the Cambridge Structural Database. Omitting this component (\textit{i.e.}, KDG) produces more broadly distributed initial conformations \cite{riniker_better_2015}.
\name{}, GeoDiff and MCF are trained from scratch on the same data. For all methods, the input is the extracted cycles without substituents and they are evaluated across the same five random data splits, with averaged results reported to ensure statistical robustness. 
For each ring, we generate twice the amount of conformers present in the test set, and assess their quality using two established metrics: Average Minimum RMSD (AMR) and coverage (see SI~\ref{SI:benchmarking_methodology} for definitions) \cite{ganea_geomol_2021}. As we generate the same number of conformers for each method, the sampling efficiency of the methods can be directly compared. AMR measures geometric performance by computing the minimum pairwise RMSD between the generated and reference conformer ensembles, and coverage quantifies the fraction of conformers successfully recovered within a 0.1 \AA{} RMSD threshold. We adopt this stringent cutoff to reflect the small size of the assessed ring systems \cite{braun_understanding_2024}.
Ideally, a conformer generator samples the entire reference conformer space while producing geometries that closely match the reference. More formally, we evaluate this by measuring both precision, \textit{i.e.}, how closely (in RMSD) the reference conformers are reproduced by the model, and recall, \textit{i.e.}, how completely the generated set spans the reference conformer space \cite{ganea_geomol_2021}.

\begin{table}[h]
  \centering
  \small 
  \begin{tabular}{lcccccccc}
    \multirow{3}{*} & \multicolumn{4}{c}{(a) Unrelaxed} & \multicolumn{4}{c}{(b) Relaxed with MMFF} \\
    \cmidrule(lr){2-5} \cmidrule(lr){6-9}
    & \multicolumn{2}{c}{Precision} & \multicolumn{2}{c}{Recall} & \multicolumn{2}{c}{Precision} & \multicolumn{2}{c}{Recall} \\
    \cmidrule(lr){2-3} \cmidrule(lr){4-5} \cmidrule(lr){6-7} \cmidrule(lr){8-9}
    Method & \shortstack{AMR\\(\AA) $\downarrow$} & \shortstack{Coverage\\(\%) $\uparrow$} & \shortstack{AMR \\ (\AA)$ \downarrow$} & \shortstack{Coverage\\(\%) $\uparrow$} & \shortstack{AMR\\(\AA) $\downarrow$} & \shortstack{Coverage\\(\%) $\uparrow$} & \shortstack{AMR \\ (\AA) $\downarrow$} & \shortstack{Coverage\\(\%) $\uparrow$} \\
    \cmidrule(l){1-1} \cmidrule(lr){2-5} \cmidrule(lr){6-9}
    \textbf{\name{} (ours)}    & \textbf{0.13} & \textbf{67.5} & \textbf{0.09} & \textbf{75.8} & \textbf{0.13} & \textbf{70.8} & \textbf{0.11} & \textbf{72.2} \\
    RDKit ETKDG (Small Cycles) & 0.17 & \underline{51.4} & 0.13 & \underline{60.1} & 0.15 & 63.4 & \underline{0.12} & 65.5 \\
    RDKit ETKDG                & 0.18 & 43.6 & 0.14 & 53.3 & 0.15 & 61.0 & 0.13 & 63.7 \\
    RDKit KDG                  & 0.20 & 33.0 & 0.15 & 51.4 & 0.15 & 60.3 & 0.13 & 64.0 \\
    MCF                        & \underline{0.16} & 46.2 & \underline{0.12} & \underline{60.0} & \underline{0.14} & \underline{64.3} & \textbf{0.11} & \underline{68.1} \\
    GeoDiff                    & 0.24 & 24.9 & 0.18 & 42.4 & 0.17 & 57.2 & 0.14 & 61.4 \\
    \cmidrule(l){1-1} \cmidrule(lr){2-5} \cmidrule(lr){6-9}
  \end{tabular}
  \vspace{2px}
  \caption{\centering{Quantitative benchmarking on the ring puckering, \textit{i.e.}, the atomic displacements from the mean plane (see Fig.~\ref{fig:methods_model_overview}a). We report the RMSD precision and recall for (a) unrelaxed and (b) relaxed structures in terms of Average Minimum RMSD (AMR) and coverage (0.1~\AA{} threshold). The best performing model is highlighted in bold, the second best underlined.}}
  \label{tab:results_rmse_random_split}
\end{table}

\begin{table}[h]
  \centering
  \small 
  \begin{tabular}{lcccccccc}
    \multirow{3}{*} & \multicolumn{4}{c}{(a) Unrelaxed} & \multicolumn{4}{c}{(b) Relaxed with MMFF} \\
    \cmidrule(lr){2-5} \cmidrule(lr){6-9}
    & \multicolumn{2}{c}{Precision} & \multicolumn{2}{c}{Recall} & \multicolumn{2}{c}{Precision} & \multicolumn{2}{c}{Recall} \\
    \cmidrule(lr){2-3} \cmidrule(lr){4-5} \cmidrule(lr){6-7} \cmidrule(lr){8-9}
    Method & \shortstack{AMR\\(\AA) $\downarrow$} & \shortstack{Coverage\\(\%) $\uparrow$} & \shortstack{AMR \\ (\AA)$ \downarrow$} & \shortstack{Coverage\\(\%) $\uparrow$} & \shortstack{AMR\\(\AA) $\downarrow$} & \shortstack{Coverage\\(\%) $\uparrow$} & \shortstack{AMR \\ (\AA) $\downarrow$} & \shortstack{Coverage\\(\%) $\uparrow$} \\
    \cmidrule(l){1-1} \cmidrule(lr){2-5} \cmidrule(lr){6-9}
    \textbf{\name{} (ours)}    & \textbf{0.18} & \textbf{47.4} & \textbf{0.15} & \textbf{51.0} & \textbf{0.16} & \textbf{61.5} & \textbf{0.14} & \textbf{61.0} \\
    RDKit ETKDG (Small Cycles) & \underline{0.22} & \underline{37.7} & \underline{0.17} & \underline{44.5} & \underline{0.17} & 55.8 & \textbf{0.14} & 56.3 \\
    RDKit ETKDG                & \underline{0.22} & 35.6 & \underline{0.17} & 42.9 & 0.18 & 54.2 & \underline{0.15} & 54.6 \\
    RDKit KDG                  & 0.23 & 26.4 & 0.18 & 41.5 & 0.18 & 54.3 & 0.16 & 55.7 \\
    MCF                        & 0.23 & 24.6 & 0.19 & 35.6 & \underline{0.17} & \underline{56.4} & \textbf{0.14} & \underline{58.2} \\
    GeoDiff                    & 0.28 & 16.4 & 0.22 & 30.5 & 0.20 & 50.9 & 0.17 & 53.9 \\
    \cmidrule(l){1-1} \cmidrule(lr){2-5} \cmidrule(lr){6-9}
  \end{tabular}
  \vspace{2px}
  \caption{\centering{Quantitative benchmarking on the atomic position RMSD. We report precision and recall for (a) unrelaxed and (b) relaxed structures in terms of Average Minimum RMSD (AMR) and coverage (0.1~\AA{} threshold). The best performing model is highlighted in bold, the second best underlined.}}
  \label{tab:results_rmsd_random_split}
\end{table}             
We now evaluate \name{} against existing methods for molecular conformer generation. Our analysis yields four key findings: 
\name{} (a) outperforms prior methods across nearly all quantitative metrics, with particularly strong improvements over Euclidean generative models (see Sec. \ref{sec:results_benchmarking}), and (b) exhibits strong performance also after force field relaxation, as commonly applied in practice (see Sec. \ref{sec:results_benchmarking}). We further (c) illustrate the capabilities of \name{} for sampling the conformer space of ring systems relevant to drug discovery and catalysis (see Sec. \ref{sec:res_examples}), and (d) show that our approach can be extended to incorporate not only the conformers of rings, but also their exocyclic substituents, for example through integration of \name{} with torsional diffusion \cite{jing_torsional_2022} (see Sec. \ref{sec:exocyclic_subst}).

\subsection{\name{} outperforms existing methods for conformer generation }\label{sec:results_benchmarking}

We benchmark the models on two quantitative measures: puckering RMSD, which measures the ability of the methods to reproduce the characteristic ring puckering by quantifying the atomic displacements from the mean plane; and all-atom RMSD, which captures the global geometric deviation across the ring atoms. Results for both measures are shown in Tabs. \ref{tab:results_rmse_random_split}a and b, with formal definitions provided in SI~\ref{SI:benchmarking_methodology}.
Standard errors quantifying the precision of the mean performance across the different data splits are reported for all methods in SI~\ref{SI:sems}. Across all methods and metrics, the standard erros over different data splits remain relatively small (0.008--0.018~\AA{} for AMR and 0.6--3.9\% for coverage), and the performance gaps are generally larger than one standard error, indicating that the reported differences are robust.

On puckering performance (see Tab.~\ref{tab:results_rmse_random_split}a), \name{} substantially outperforms Euclidean models in cyclic conformer generation. Compared to MCF, which is currently the model architecture that achieves state-of-the-art results on GEOM-Drugs, \name{} achieves higher coverage precision (67.5$\pm$1.2\% vs.~46.2$\pm $1.6\%) and recall (75.8$\pm $0.6\% vs.~60.0$\pm $1.2\%). The improvements are even more pronounced compared to GeoDiff, with \name{} for example reducing the mean AMR from 0.24$\pm $0.016~\AA{} to 0.13$\pm$0.008~\AA{} (precision) and from 0.18$\pm$0.018~\AA{} to 0.09$\pm$0.008~\AA{} (recall). RDKit performs comparably to MCF and worse than \name{} on puckering performance.
For all-atom RMSD (see Tab.~\ref{tab:results_rmsd_random_split}a), \name{} achieves higher precision than competing methods, indicating that generated conformers more closely match reference structures, while exhibiting similar or slightly improved recall, reflecting comparable coverage of conformer diversity. Compared to MCF, \name{} reduces the mean AMR from 0.23$\pm$0.015~\AA{} to 0.18$\pm$0.009~\AA{} in precision and from 0.19$\pm$0.013~\AA{} to 0.15$\pm$0.010\AA{} in recall. RDKit shows reduced precision, highlighting the limited fidelity of rule-based torsional sampling, but achieves recall comparable to \name{}, likely due to its broad initialization strategy. Overall, \name{} generates conformers that are both geometrically precise and structurally diverse, faithfully capturing the essential puckering modes of cyclic molecules.

In practice, generated conformers are almost always relaxed with classical or machine learning-based force fields. Tabs. ~\ref{tab:results_rmse_random_split}b and ~\ref{tab:results_rmsd_random_split}b show that after MMFF94 ~\cite{halgren_merck_1996} relaxation, \name{} shows strong precision and competitive recall compared to other methods, despite using substantially fewer parameters (966K) than other machine learning models (\textit{e.g.,} MCF: 13M). In the results for \name{} presented here, we use 30 inference steps. In SI~\ref{SI:inference_step_count}, we demonstrate that as few as 2--5 inference steps are sufficient to achieve high-quality samples, which enables efficient large-scale conformer generation. In comparison, the usual number of generation steps for GeoDiff and MCF are 5000 and 50, respectively.

\subsection{\name{} captures ring conformer distributions}\label{sec:res_examples}
\begin{figure}[h]
    \centering
    \includegraphics[width=\textwidth]{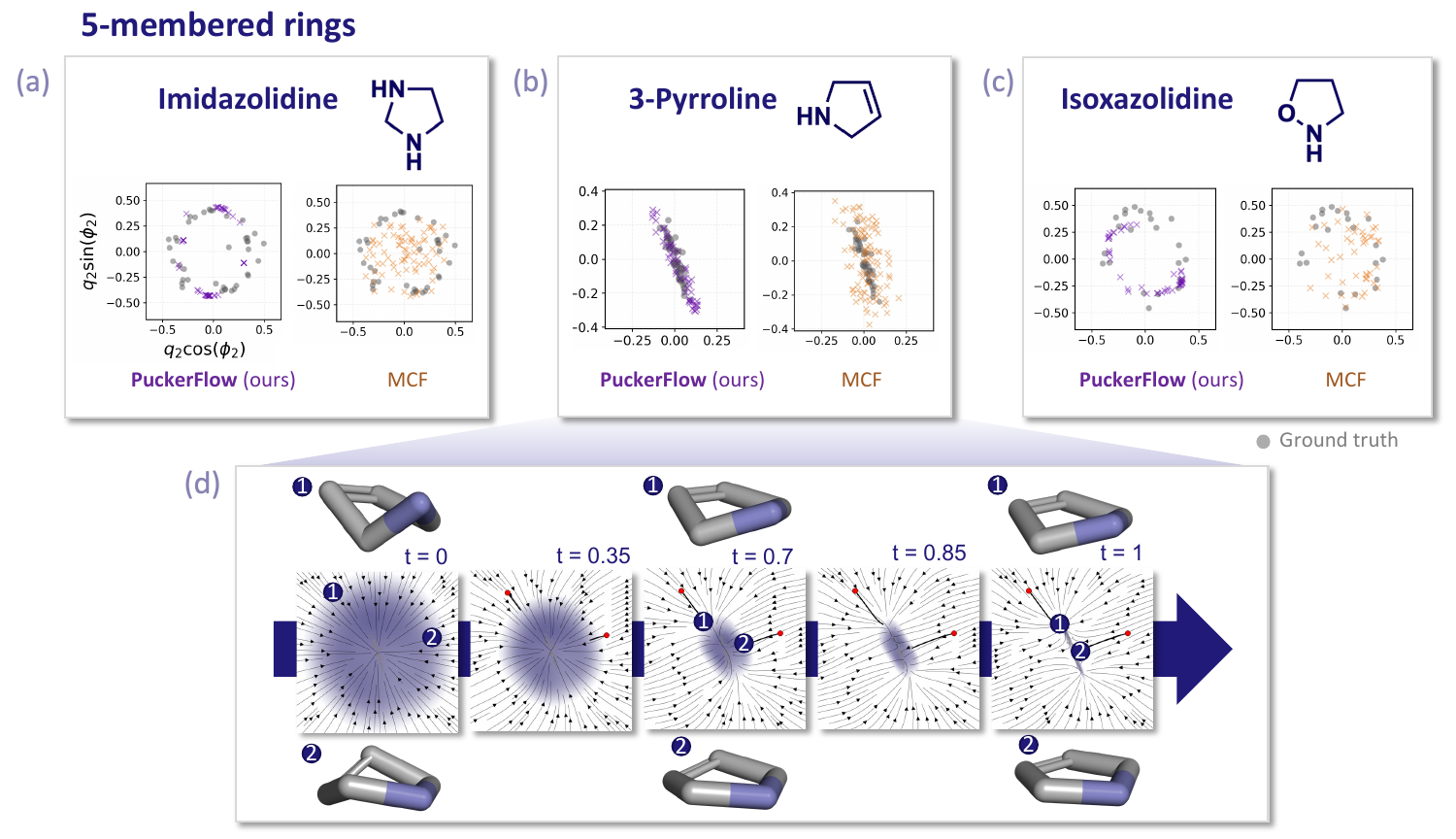}
    \caption{(a-c) Generated conformer space of five-membered rings with \name{} (violet) and MCF (gold) visualized in the two-dimensional Cremer-Pople space. Grey circles denote ground-truth conformations. (d) Visualization of the vector field and sampling distribution during inference for the imidazolidine ring: prior distribution (left), intermediate timesteps (middle), and final generated distribution (right). Highlighted points correspond to structures along the inference trajectory, which all correspond to valid, closed rings.}
    \label{fig:Results_5}
    \end{figure}

\begin{figure}[t]
    \centering
    \includegraphics[width=\textwidth]{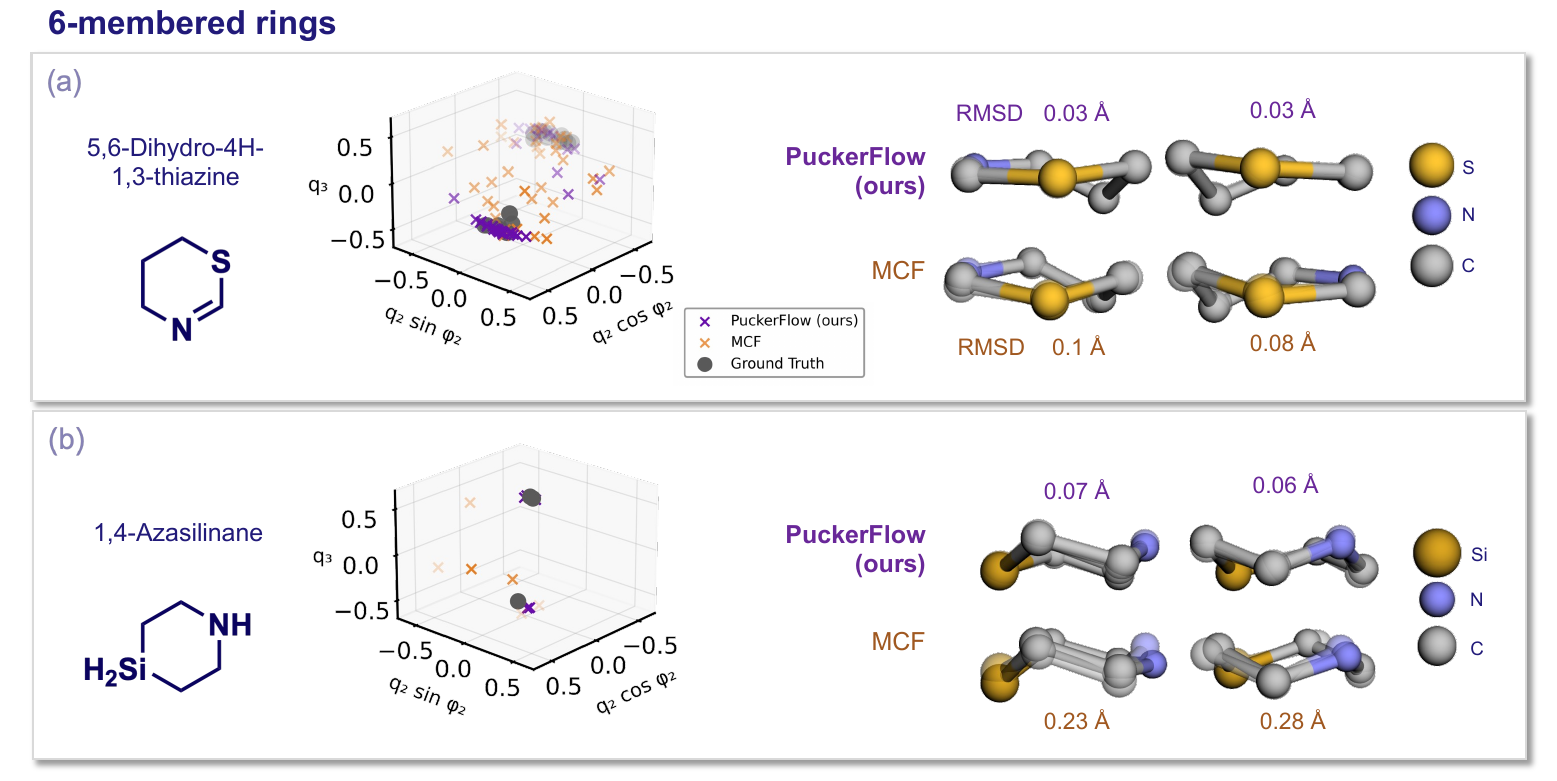}
    \caption{(a-b) Generated conformers of six-membered rings obtained with \name{} and MCF. Left: generated conformers shown in Cremer-Pople space as purple (\name{}) and gold (MCF) crosses, with gray circles denoting ground-truth conformers. Right: representative ground-truth conformers (transparent) and their closest generated counterparts.}
    \label{fig:Results_6}
    \end{figure}

The low dimensionality of the Cremer-Pople coordinates enables intuitive visualization of the conformer ensembles: five- and six-membered rings can be represented in two and three dimensions, respectively. Figs. \ref{fig:Results_5} and \ref{fig:Results_6} show the distribution of generated test-set rings in Cremer-Pople space, spanning diverse bonding patterns, heteroatoms, and ring sizes.
For each system, we identify representative ground-truth conformers through $k$-means clustering in Cremer-Pople space and compare them with the closest structures generated by \name{} and MCF, the best-performing Euclidean generative model as identified in the preceding section. 

Among the five-membered rings, the conformers of imidazolidine exhibits nearly constant amplitude with varying phase (see Fig.~\ref{fig:Results_5}a). Here, \name{} covers most of the angular phase space and closely reproduces the constant amplitude, while MCF samples a wide range of amplitudes (see Fig.~\ref{fig:Results_5}a--c). In contrast, 3-pyrroline --- a substructure relevant to antioxidants~\cite{nguyen_synthesis_2022} and drug candidates, \textit{e.g.}, for myotonic syndrome~\cite{de_bellis_dual_2018} --- exhibits a puckering mode at a fixed phase angle with variable amplitude (see Fig.~\ref{fig:Results_5}b), which is correctly reproduced by \name{} but shows an overly wide phase spread for MCF.
We observe similar results for isoxazolidine (Fig.~\ref{fig:Results_5}c), an important scaffold in natural products, drug discovery~\cite{berthet_isoxazolidine_2016} and organocatalysis~\cite{doyle_nh-isoxazolo-bicycles_2011}.
Fig.~\ref{fig:Results_5}d visualizes the vector field and sampling distribution in Cremer-Pople coordinates during generation, illustrating that \name{} produces closed valid rings along the whole inference trajectory.

For six-membered rings, we illustrate the three-dimensional Cremer-Pople distributions of 5,6-dihydro-4H-1,3-thiazine and 1,4-azasilinane in Fig.~\ref{fig:Results_6}. The former system shows envelope conformers, while the
latter, previously studied in the general context of sila-heterocyclohexanes~\cite{shainyan_silacyclohexanes_2020}, exhibits two chair conformers (Fig. \ref{fig:Results_6}b). \name{} correctly identifies both conformer modes in Cremer-Pople space, while MCF does not recover this bimodal distribution.
For larger rings, the higher dimensionality of Cremer-Pople space complicates visualization; therefore, these systems are here presented in Euclidean space. Fig.~\ref{fig:Results_7_8}a highlights the conformer space of the medium-sized systems: the seven-membered 1,4-oxazepane and 1,3-diazepine and the eight-membered 1,3,6,2-dioxaselenaphosphocane. These rings have been used in a variety of applications \cite{li_conformations_2001, faisca_phillips_modern_2020}, with for example 1,4-oxazepane being an essential substructure in several clinically used drugs, \textit{e.g.}, brensocatib for bronchiectasis \cite{chalmers_phase_25}. 
Across these challenging examples, \name{} shows strong performance compared to MCF.

\subsection{Generating exocyclic substituents}\label{sec:exocyclic_subst}
\begin{figure}[t!]
    \centering
    \includegraphics[width=\textwidth]{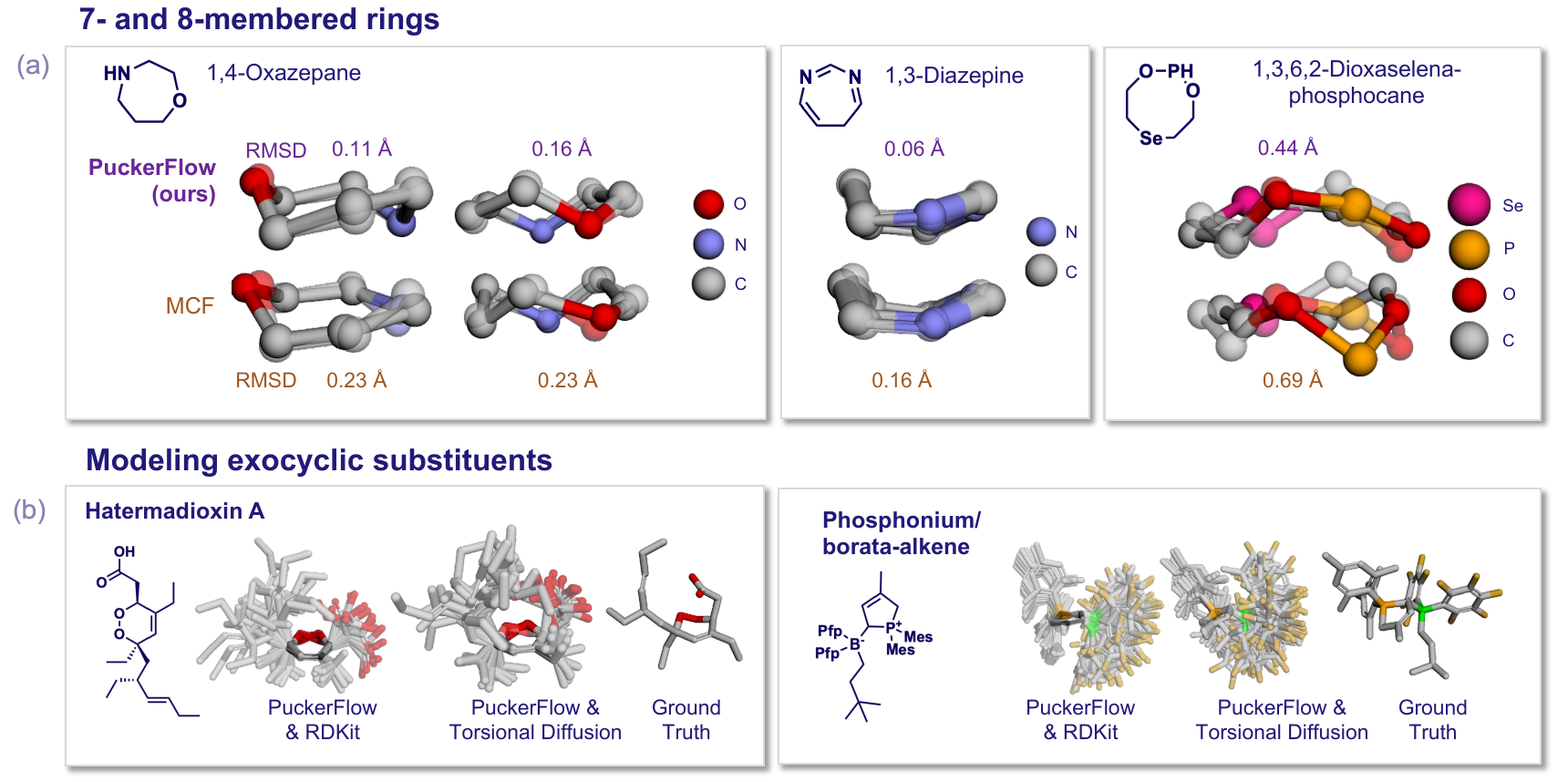}
    \caption{(a) Generated conformers of seven- and eight-membered rings from \name{} and MCF. (b) \name{} can be integrated into RDKit's ETKDG or torsional diffusion to model exocyclic substituents. (Mes = mesityl, Pfp = pentafluorophenyl) }
    \label{fig:Results_7_8}
    \end{figure}

While \name{} generates cyclic substructures, it allows for complete molecular conformer generation when combined with established methods that model exocyclic substituents, such as RDKit's ETKDG \cite{landrum_rdkitrdkit_2024} or torsional diffusion \cite{jing_torsional_2022}. Fig.~\ref{fig:Results_7_8}b demonstrates this workflow, where \name{}-generated ring conformers serve as initial local substructures for these approaches to generate the coordinates of the exocyclic substituents (see SI \ref{SI:exocyclic} for details). 
We illustrate this approach on two chemically diverse systems: hatermadioxin A, a natural product exhibiting cytotoxicity against multiple human cancer cell lines~\cite{takada_isolation_2001}, and a zwitterionic phosphonium/borata-alkene structure synthesized via frustrated Lewis pair addition~\cite{krupski_unusual_2016}. Both molecules display significant structural flexibility in their exocyclic substituents. These results demonstrate that \name{} can enhance widely used conformer generators by providing more precise sampling of ring conformer ensembles, which then serve as realistic scaffolds for modeling full molecular geometry.

\section{Conclusion}
We have presented \name{}, a generative model that performs flow matching on the low-dimensional Cremer-Pople space to efficiently generate conformers of cyclic systems. 
To enable generation in Cremer-Pople space, we introduced a manifold-adapted cyclic Fourier filter that enables the model to handle rings of varying sizes, and derived a geometry-informed prior distribution that enables generation of closed rings. 
Evaluated on diverse ring systems, \name{} outperforms existing methods across almost all metrics, demonstrating strong performance in reliably sampling the rich conformer space of cyclic systems. We show the strengths of \name{} in sampling ring systems relevant in drug discovery and catalysis, and that \name{} retains strong performance at only few inference steps, promoting sustainable use of compute resources due to its sampling efficiency.
As \name{} exhibits particularly strong improvements over Euclidean generative models, our results highlight the potential of manifold diffusion in internal coordinate systems that enables efficient learning of the relevant degrees of freedom.

For future work, we propose three extensions of \name{}: (a) full integration into other manifold generative models, (b) training on macrocycles and (c) property-guided generation.
For (a), our method is already fully compatible with generative modeling over torsional angles \cite{jing_torsional_2022,lin_ppflow_2024}), making an extension feasible that performs joint flow matching over the different degrees of freedom to simultaneously sample the cyclic and torsional degrees of freedom.
\name{} can further be incorporated into the widely used protein-ligand model DiffDock \cite{corso_diffdock_2022} to enhance the modeling of cyclic molecules in drug discovery.
For (b), the currently used dataset derived from diverse sources predominantly features small- to medium-sized rings prominent in many areas of chemistry \cite{reyes_construction_2021}. With only 1\% of rings containing nine ring atoms or more, it lacks sufficient training data for these larger systems. Extending \name{} to macrocycles is feasible, and future work will explore handling their non-convex projection on the mean plane via discrete flow matching.
For (c), which is important for downstream applications in catalysis or drug design, \name{} can be fine-tuned for property-guided generation to, \textit{e.g.}, produce ring geometries tailored for interaction with a target substrate or protein. 
Owing to its end-to-end differentiable design, \name{} is directly compatible with gradient-based fine-tuning schemes such as adjoint matching \cite{domingo_i_enrich_adjoint_2025}.
In contrast, RDKit’s ETKDG \cite{landrum_rdkitrdkit_2024} and other generative models tailored to rings \cite{grambow_accurate_2024} are not end-to-end differentiable, preventing their use in gradient-based fine-tuning.

By enabling robust and chemically-informed generation of cyclic conformers, \name{} addresses the unique challenges of reliably modeling cyclic molecules in applications across chemistry and biology. 


\section{Acknowledgements}
We thank Vignesh Ram Somnath, Andreas Krause and Martin Diehl for constructive discussions. This publication was created as part of NCCR Catalysis (grant numbers 180544 and 225147), a National Centre of Competence in Research funded by the Swiss National Science Foundation. A. Hartgers was supported by an NCCR Catalysis Young Talents Fellowship.

\section{Code and Data Availability}
Code for \name{} is available at \url{https://github.com/digital-chemistry-laboratory/puckerflow}. The data used for training and evaluation is provided at \url{https://doi.org/10.5281/zenodo.18243479}.


\printbibliography
\end{refsection}
\savestatus
\clearpage
 \begin{refsection}

\appendix
\section{Supporting Information}\label{SI}
\subsection{Additional Background on Cremer-Pople Coordinates}\label{SI:cp_background}
This section provides background information on the Cremer-Pople coordinates. We describe the mathematical details of the interconversion between Cremer-Pople and Cartesian coordinates in SI~\ref{SI:cp_to_cart}–\ref{SI:cart_to_cp} following \cite{cremer_general_1975} and \cite{cremer_calculation_1990}. 

\subsubsection{Cartesian to Cremer-Pople coordinates}\label{SI:cart_to_cp}
Before calculating the Cremer-Pople coordinates from Cartesian coordinates, the Cartesian coordinate system must be standardized. The origin is defined as the mean position of all ring atoms,
\begin{equation}
\sum_{j=1}^N \mathbf{R}_j = 0, \label{eq:f0}
\end{equation}
where $\mathbf{R}_j$ denotes the Cartesian coordinates of atom $j$, and $N$ is the total number of atoms in the ring.
We define the two axes of the mean plane, which form the first two axes of the coordinate system, as
\begin{subequations}
 \begin{align}
    \mathbf{R'} &= \sum_{j=1}^N \mathbf{R_j} \cos[2\pi(j-1)/N] \\
    \mathbf{R''} &= \sum_{j=1}^N \mathbf{R_j} \sin[2\pi(j-1)/N].
\end{align}     
 \end{subequations}
 
\begin{figure}[hb]
    \centering
    \includegraphics[width=0.4\linewidth]{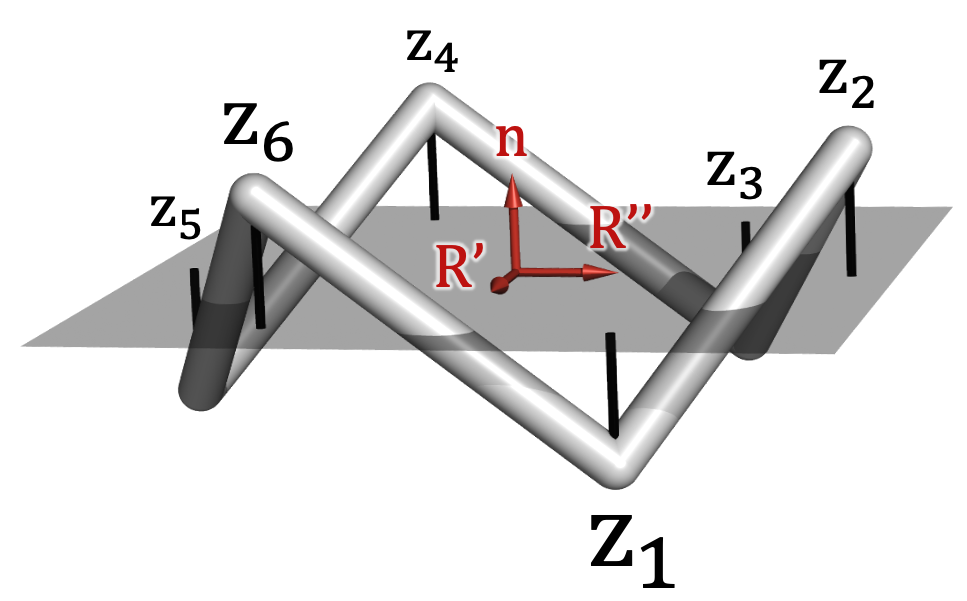}
    \caption{Standardized Cremer-Pople Cartesian coordinate system.}
    \label{fig:construction_cp}
\end{figure}

The third axis, normal to the mean plane, is then defined as
\begin{equation}
\mathbf{n} = \frac{\mathbf{R'} \times \mathbf{R''}}{|\mathbf{R'} \times \mathbf{R''}|},
\end{equation}
as illustrated in Fig.~\ref{fig:construction_cp}, from which the out-of-plane displacement of each atom follows as
\begin{equation}
z_j = \mathbf{R}_j \cdot \mathbf{n}.
\end{equation}

From these displacements $z$, the Cremer-Pople coordinates can be calculated with the formulas given in Equations~\ref{eq:cp_formula}--\ref{eq:cp_formula_odd} of the main paper. Since atom ordering affects the phase angles $\phi$, we use the canonicalization conventions given in SI~\ref{SI:numbering}.

\subsubsection{Cremer-Pople to Cartesian Coordinates}\label{SI:cp_to_cart}
\begin{figure}[h]
    \centering
    \includegraphics[width=0.5\linewidth]{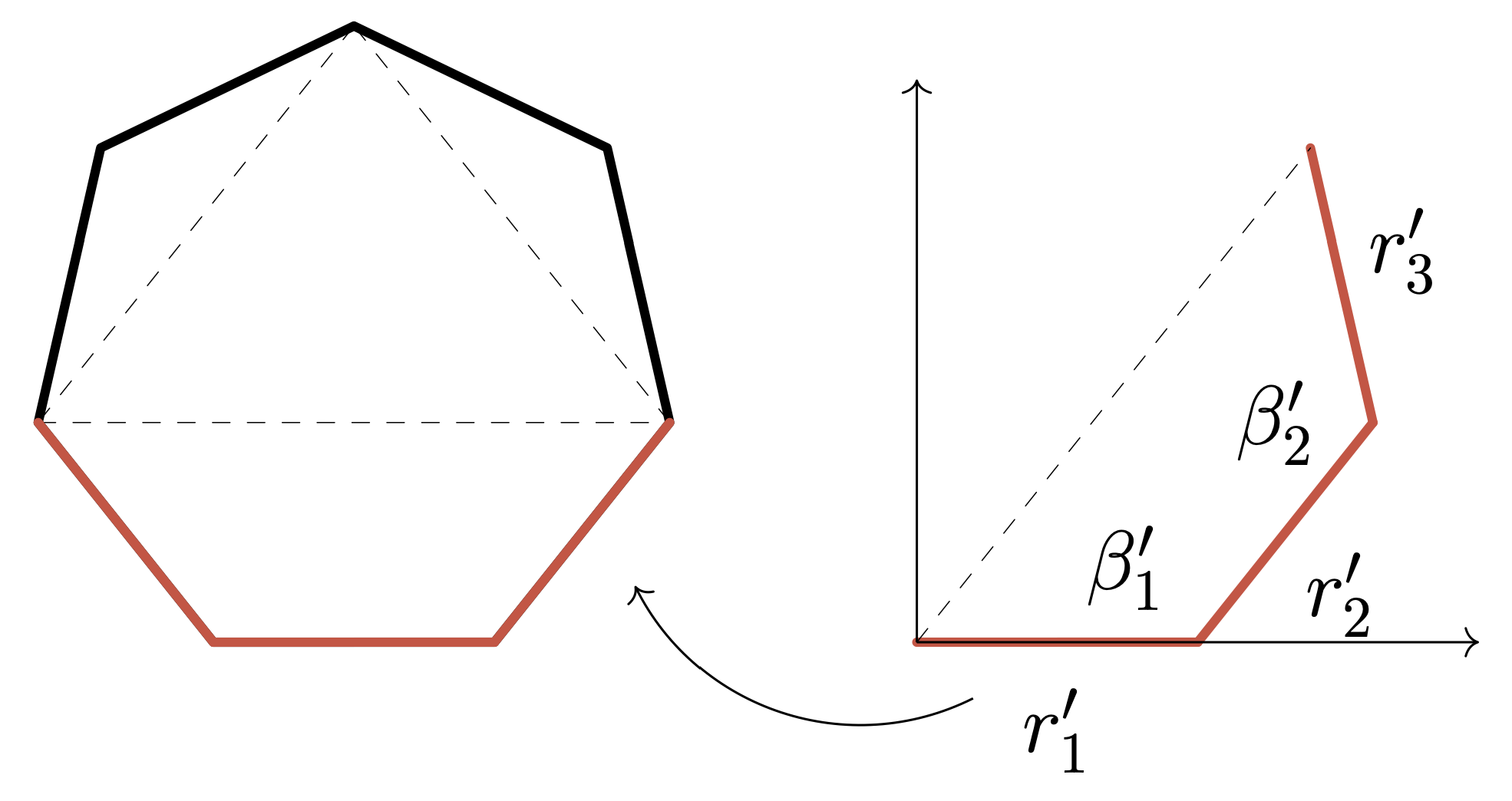}
    \caption{Reconstruction of the projected atomic coordinates on the mean plane. First, three segments are created based on the projected bond angles $\beta'$ and projected bond lengths $r'$ (right), which are subsequently assembled to form the complete ring (left). Figure adapted from Cremer \cite{cremer_calculation_1990}.}
    \label{fig:segmentation}
\end{figure}

The reconstruction of Cartesian coordinates from Cremer-Pople coordinates is performed in two steps. First, the out-of-plane displacements $z$ are obtained, followed by the determination of the $x$- and $y$-coordinates. The out-of-plane displacements $z$ can be directly obtained from the Cremer-Pople coordinates by
\begin{equation}
        z_j =(2/N)^{1/2}\sum_{m=2}^{(N-1)/2}q_m \cos(\phi_m+m\alpha_j)\
\end{equation}

if the number of atoms in the ring $N$ is odd, and   

\begin{equation}
    z_j =(2/N)^{1/2}\sum_{m=2}^{N/2-1}q_m \cos(\phi_m+m\alpha_j) + (1/N)^{1/2}\ q_{N/2}\ (-1)^{j-1}
\end{equation}

if $N$ is even, with $\alpha_j= 2 \pi (j-1) / N$. Next, the atomic coordinates are projected onto the mean plane, thereby removing the contribution of the out-of-plane displacements. This projection requires predefined bond angles $\beta_{ijk}$ (defined by consecutive atoms $i$, $j$, and $k$) and bond lengths $r_{ij}$ (between adjacent atoms $i$ and $j$). The projected variables are then obtained as follows
 
\begin{align}
    r_{ij}' &= \sqrt{r_{ij}^2-(z_j-z_i)^2} \label{eq:bondlength}\\
    \cos \beta_{ijk}' &= \frac{(z_k-z_i)^2-(z_j-z_i)^2-(z_k-z_j)^2+2r_{ij}r_{jk}\cos(\beta_{ijk})}{2r_{ij}'r_{jk}'} \label{eq:bondangle}
\end{align}
where the prime symbol ($'$) denotes the projected version of the variable, \textit{e.g.,} $r_{ij}'$ is the projected bond length between atoms $i$ and $j$. The ring is first separated into three segments as depicted in Fig.~\ref{fig:segmentation}, which are constructed separately. The length of each segment (see Tab.~\ref{tab:segmentation}), as well as the subsequent assembly of the segments into the ring follows the method introduced by Cremer~\cite{cremer_calculation_1990}.

\begin{table}[h!]
    \centering
    \renewcommand{\arraystretch}{1.1}
    \begin{tabular}{c|*{3}{p{0.3cm}}*{3}{p{0.3cm}}*{3}{p{0.3cm}}}
        & \multicolumn{3}{c}{\# Atoms in Seg.} & \multicolumn{3}{c}{\# Bonds in Seg.} & \multicolumn{3}{c}{\# Angles in Seg.} \\
        \hline
        $N$ & $S_1$ & $S_2$ & $S_3$ & $S_1$ & $S_2$ & $S_3$ & $S_1$ & $S_2$ & $S_3$ \\
        \hline
        5 & 3 & 2 & 3 & 2 & 1 & 2 & 1 & 0 & 1 \\
        6 & 3 & 3 & 3 & 2 & 2 & 2 & 1 & 1 & 1 \\
        7 & 3 & 4 & 3 & 2 & 3 & 2 & 1 & 2 & 1 \\
        8 & 4 & 3 & 4 & 3 & 2 & 3 & 2 & 1 & 2 \\
    \end{tabular}
    \vspace{0.5em}
    \caption{Partitioning of the projected ring into three segments (\(S_1\), \(S_2\), and \(S_3\)), with the number of atoms, bonds, and angles in each segment specified for \(N\)-membered rings ranging from sizes 5 to 8. Table adapted from Cremer \cite{cremer_calculation_1990}.}
\label{tab:segmentation}
\end{table}

\subsection{Computational Details on Cremer-Pople coordinates}\label{SI:cp_computational_details}
The Cremer-Pople coordinates represent the \textit{puckering} of a ring in a compressed manner. However, this compression imposes constraints on the internal parameters. Prior research has primarily employed Cremer-Pople coordinates to analyze existing, and therefore geometrically feasible, molecules \cite{chan_understanding_2021}. In contrast, we apply these coordinates in a generative context. This application requires several computational and methodological considerations to be addressed, including geometric feasibility (see SI~\ref{SI:cp_feasibility}) and atom numbering conventions (see SI~\ref{SI:numbering}).

\subsubsection{Feasibility}\label{SI:cp_feasibility} 
Cremer-Pople coordinates describe the displacements $z$, which, combined with a set of imposed bond lengths $r$ and angles $\beta$, are used to reconstruct a ring's three-dimensional structure. These restrictions are imposed by the conversion equations (see SI~\ref{SI:cp_to_cart}), which are not defined for all possible input parameters.
The first restriction involves the projected bond length $r_{ij}'$: 
\begin{equation} r_{ij}' = \sqrt{r_{ij}^2-(z_j-z_i)^2}. \tag{\ref{eq:bondlength}} \end{equation} Equation~\ref{eq:bondlength} requires the term under the square root to be non-negative. This imposes the first condition: the absolute difference between consecutive displacements cannot exceed the bond length, \textit{i.e.}, $|z_{j}-z_{i}| \leq r_{ij}$. We mitigate this problem by constraining the amplitude of the prior distribution, thereby ensuring feasible puckering amplitudes (see SI~\ref{SI:prior_distribution} for details on the prior). The second restriction involves the projected bond angle $\beta_{ijk}'$
 : \begin{equation} \cos \beta_{ijk}' = \frac{(z_k-z_i)^2-(z_j-z_i)^2-(z_k-z_j)^2+2r_{ij}r_{jk}\cos(\beta_{ijk})}{2r_{ij}'r_{jk}'}. \tag{\ref{eq:bondangle}} \end{equation} Equation~\ref{eq:bondangle} defines the cosine of the projected angle in terms of three displacements $z$, two bond lengths $r$, and the bond angle $\beta$. This introduces a second constraint: the value must lie within the valid range for a cosine,\textit{ i.e., }$[-1, 1]$. We enforce this by clipping any out-of-range values to the nearest bound.

A third restriction arises from the ring segmentation procedure. The reconstruction procedure assumes that projected angles are always added convexly, which is valid for small- and medium-sized rings. To ensure reliable performance, we therefore restrict our analysis to rings with eight or fewer members. No concave rings were observed among seven-membered rings. Among eight-membered conformers, 12 out of 104 were concave, which was considered acceptable. Nine-membered and larger rings were excluded due to the limited number of training data available (fewer than 1\% of conformers) and the increased likelihood of concavity.

\subsubsection{Atom Numbering}\label{SI:numbering}
The numbering of atoms within a ring determines the phase $\phi$ of the Cremer-Pople coordinates. In this work, we adopt the canonicalization scheme proposed by Chan et al. \cite{chan_understanding_2021}, which establishes a unique numbering by comparing sequences of atom and bond types. The rules are applied hierarchically:
\begin{enumerate}
\item \textbf{Bond Order:} Higher bond orders take precedence over lower ones. \par
    \hspace*{6em}\textit{i.e.}, triple > double > single
\item \textbf{Atomic Number:} Lower atomic numbers take precedence over higher numbers.
\end{enumerate}
These rules are applied cumulatively to determine the canonical starting atom and numbering direction. If a unique ordering cannot be determined from a single atom or bond, adjacent atoms and bonds are considered incrementally until the entire ring is included.
This procedure ensures an unambiguous atom and bond pattern. Note that these are all bond orders present in the dataset, as aromatic rings are excluded (see SI~\ref{SI:data_preprocessing}).

\subsection{Data Preprocessing}\label{SI:data_preprocessing}
To prepare the data for \name{}, the dataset by Folmsbee et al. \cite{folmsbee_systematic_2023} is preprocessed through three steps: (1) ring extraction, (2) bond parameter dictionary creation (bond lengths and angles), and (3) a reconstructability check. Here, a \emph{ring} is defined as a sequence of covalently bonded atoms forming a single closed loop (monocyclic), excluding any substituents attached to the ring atoms. During ring extraction, ring substituents are replaced with hydrogen atoms. If the substitution procedure does not preserve hybridization, the ring is excluded from the dataset. An example is the replacement of a carbonyl (C=O) double bond with two C–H single bonds, which would change the carbon's hybridization state. Additionally, a cutoff of 1,000 conformers per ring is applied and aromatic rings are excluded, since their (limited) distortions from planarity are predominantly induced by ring fusion or exocyclic atoms rather than intrinsic puckering. The Cremer-Pople coordinates for the conformers are computed using the Python ring-puckering library developed by Chan et al. \cite{chan_understanding_2021}, which is also employed to convert between Cremer-Pople and Cartesian coordinates.

For each training set of the five splits (see main text), a dictionary of bond lengths and angles is generated to enable the reconstruction of molecular rings from Cremer-Pople coordinates. Each dictionary entry is indexed by a key that specifies the local bonding pattern and ring size in terms of atomic number ($Z$), bond order ($b$) and ring size ($r$). Bond length keys take the form $(Z_1,b,Z_2,r)$, while bond angle keys follow $(Z_1,b_1,Z_2,b_2,Z_3,r)$. The stored value for each key is the mean bond length and angle, respectively, calculated on all occurrences of that pattern in the training data. If a queried pattern is absent in the training data, the closest available pattern is selected using an empirically defined weighted distance metric. Differences in atom type (represented by atomic number $Z$) are weighted three times more heavily than differences in bond type $b$ (represented as numerical values, \textit{e.g.}, double = 2.0) and ring size $r$. The distance between two keys is therefore defined as
\begin{align}
\text{distance} = \sum  |\Delta b| + 3 \sum |\Delta Z| + 3 |\Delta r|,
\end{align}
where $\Delta$ denotes the difference between corresponding atomic numbers $Z$, bond orders $b$, and ring size $r$ of the two keys. If keys are equally close, the metric resolves the tie by selecting the larger numerical value. For example, suppose we need the bond length for a single (1.0) O–S ($Z=8 \text{ and } 16$) bond in a six-membered ring (6). Since atom–bond–atom patterns can be read in either direction (8, 16 or 16, 8), we use the lexicographically smaller key $(8,1.0,16,6)$ as the canonical form. 

After this step, we verify whether the molecules can be successfully reconstructed from the Cremer-Pople coordinates and the generated bond dictionary, as not all Cremer-Pople coordinates lie on the feasible manifold (see SI~\ref{SI:cp_feasibility}). Even if computed from feasible molecules, the lookup of bond lengths and angles can lead to a geometrically infeasible reconstruction, as detailed in SI~\ref{SI:cp_feasibility}. Conformers that cannot be reconstructed (only $\sim~0.01 \%$ of the total ring structures) are discarded from all splits.

\subsubsection{Dataset Statistics}\label{SI:data_stats}
After preprocessing, a total of 15,204 rings were retained.
The composition of this dataset is summarized in Tabs.~\ref{tab:dataset_source} and \ref{tab:dataset_ringsize}. Tab.~\ref{tab:dataset_source} provides a breakdown by data source, showing that the majority of structures originate from the Crystallography Open Database (COD, 8,996) \cite{vaitkus_workflow_2023, merkys_graph_2023, vaitkus_validation_2021, quiros_using_2018, merkys_it_2016, grazulis_computing_2015, grazulis_crystallography_2012, grazulis_crystallography_2009, downs_american_2003} and Pitt Quantum Repository (PQR, 3,169) \cite{re3dataorg_pitt_2020} databases. Tab.~\ref{tab:dataset_ringsize} reports the structural distribution, indicating that the dataset is dominated by five-membered (6,725 conformers, 141 unique) and six-membered (8,005 conformers, 209 unique) rings. 

\begin{table}[ht]
    \centering
    \begin{minipage}{0.48\textwidth}
        \centering
        \begin{tabular}{lll}
            Dataset & \# Conformers&\# Rings\\
            \hline
            COD & 8996 &361\\
            PQR& 3169  &169\\
            ZINC \cite{irwin_zinc_2005} & 2004  &36\\
            Platinum \cite{pires_platinum_2015} &  1035 &40\\
            \hline
            Total & 15 204  & 419\\
        \end{tabular}
        \caption{Dataset composition by source.}
        \label{tab:dataset_source}
    \end{minipage}
    \hfill
    \begin{minipage}{0.48\textwidth}
        \centering
        \begin{tabular}{lll}
            Ring size & \# Conformers & \# Rings\\
            \hline
            5 &  6725 & 141\\
            6 &  8005 & 209\\
            7 &  370 & 46\\
            8 &  104 & 23\\
            \hline
            Total & 15 204 & 419 \\
        \end{tabular}
        \caption{Dataset composition by ring size.}
        \label{tab:dataset_ringsize}
    \end{minipage}
\end{table}

To verify that the molecular reconstruction with dictionary-based bond parameters produces geometrically accurate molecules, we compared the dictionary values against the actual bond parameters in the validation set. The dictionary shows good consistency with the true geometries: for bond angles, the median absolute difference is 0.85$\degree$ (mean 1.96$\degree$), and for bond lengths, the median absolute difference is 0.008 \AA{} (mean 0.015 \AA{}). The signed error distributions are centered around zero, indicating no systematic bias.

\begin{figure}[h]
    \centering
    \includegraphics[width=0.4\linewidth]{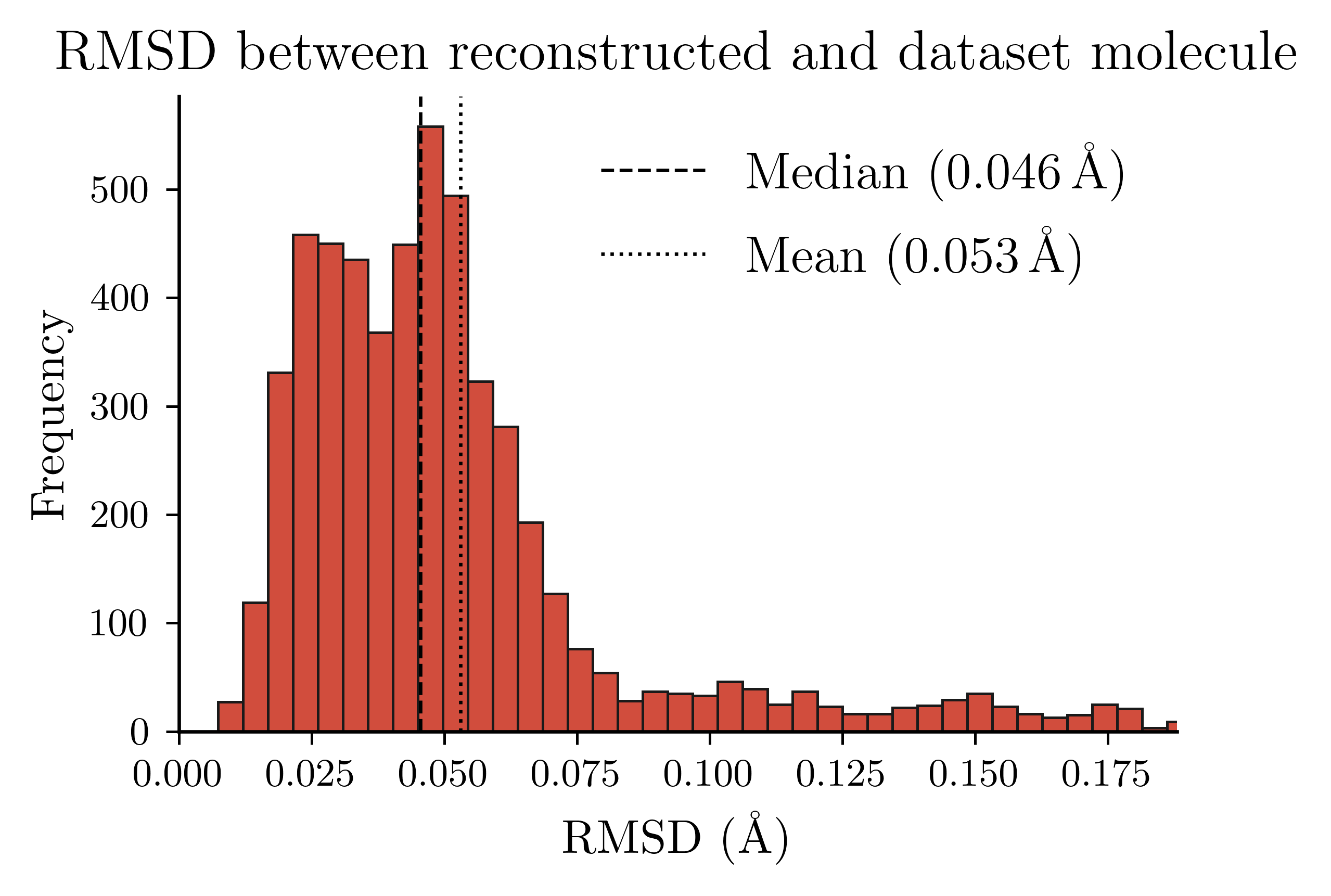}
    \caption{Distribution of the RMSD between the original molecular geometries and geometries reconstructed from the true Cremer-Pople coordinates.}
    \label{fig:rmsd_distribution}
\end{figure}

Beyond the individual bond parameters, we also assessed the overall geometric accuracy of the reconstruction procedure by evaluating the root-mean-square deviation (RMSD) between the original and reconstructed molecular geometries. The distribution of RMSD values is shown in Fig.~\ref{fig:rmsd_distribution}. The median RMSD of 0.046 \AA{} confirms that the reconstructed geometries closely reproduce the reference structures, thereby validating this approach.

\subsection{Generative Modeling}\label{SI:generative_modeling}
\subsubsection{Flow Matching}\label{SI:flow_matching}
Flow Matching (FM), introduced by Lipman et al. \cite{lipman_flow_2022}, is a generative modeling paradigm that combines techniques from diffusion models \cite{ho_denoising_2020} and \textit{Continuous Normalizing Flows} \cite{chen_neural_2018}. In this framework, a parametric vector field $u_\theta : \mathbb{R}_+ \times \mathbb{R}^d \rightarrow \mathbb{R}^d
$ is learned. This vector field defines a continuous probability path (a \textit{flow}) that transforms the prior distribution $p_0$ into the modeled distribution $p_1$, intended to match the target data distribution $q_1$. This flow is governed by the continuity equation, which constitutes an Ordinary Differential Equation (ODE) for the log probability \cite{mathieu2024flow}:
\begin{equation}
    \log p_1(x) = \log p_0(x_0) - \int_0^1 (\nabla \cdot u_\theta)(x_t) \, dt.
\label{eq:ode}
\end{equation}
The parametric vector field is then learned through the following regression objective for $u_\theta$
\begin{equation}
    \mathcal{L}_{FM}(\theta) = \mathbb{E}_{t \sim \mathcal{U}[0,1],x \sim p_t} \left[ \| u_\theta(t, x) - u(t, x) \|^2 \right].
\end{equation}
However, the true marginal vector field $u$ is not known. To address this, Flow Matching relies on the fact that the gradient of the marginal loss $\mathcal{L}_{FM}$ is equal to the gradient of the Conditional Flow Matching (CFM) loss ($\nabla\mathcal{L}_{CFM}=\nabla\mathcal{L}_{FM}$). CFM transforms the objective by marginalizing over the data samples $x_1\sim q_1$, leading to the loss function based on the conditional vector field $u_t(x|x_1)$
\begin{equation}
    \mathcal{L}_{\text{CFM}}(\theta) = \mathbb{E}_{t \sim \mathcal{U}[0,1], x_1 \sim q, x_t \sim p_t(x | x_1)} \left[ \| u_\theta(t, x) - u_t(x \mid x_1) \|^2 \right]
\end{equation}

which can be used to train $u_\theta$. Once trained, sampling from the target distribution is achieved by integrating the learned ODE (Equation~\ref{eq:ode}). Flow Matching's free choice of prior distribution, a capability absent in diffusion models, is key to overcoming the feasibility constraints imposed by the Cremer-Pople manifold (see SI \ref{SI:cp_feasibility}).

\subsubsection{Prior Distribution}\label{SI:prior_distribution}
The manifold of feasible Cremer-Pople coordinates depends on the structure-dependent bond parameters of the ring. Consequently, the maximum permissible Cremer-Pople amplitudes can vary between compounds. In practice, we sample uniformly in Cremer-Pople space, restricting the maximum amplitudes to 0.8, 0.56, and 0.4 for coordinates of order 
$m=2,3, \text{and } 4$, respectively, with the decrease in amplitude originating in the normalization of the Cremer-Pople coordinates.

\subsubsection{Pseudocode}\label{SI:pseudocode}
This section presents the pseudocode for training and inference, shown in Algorithms~\ref{alg:flow_training} and~\ref{alg:flow_inference}, respectively.

\begin{algorithm2e}[H]
\caption{Flow Matching training with Cremer-Pople coordinate ($\mathbf{x}_1$) prediction}\label{alg:flow_training}
\KwIn{Training data distribution \( p_1 \), prior \( p_0 \), initialized vector field \( u_\theta \)}
\While{$\text{\upshape Training}$}{
    \( \mathbf{x}_0 \sim p_0(\mathbf{x}_0) \); \( \mathbf{x}_1 \sim p_1(\mathbf{x}_1) \); \( t \sim \mathcal{U}(0, 1) \)\;
    \( \mathbf{x}_t \leftarrow t \mathbf{x}_1 + (1 - t) \mathbf{x}_0 \)\;
    \( \mathcal{L}_{\text{CFM}} \leftarrow \|u_\theta(\mathbf{x}_t, t) - \mathbf{x}_1\|^2 \)\;
    \( \theta \leftarrow \text{Update}(\theta, \nabla_\theta \mathcal{L}_{\text{CFM}}) \)\;
}
\Return \( u_\theta \)
\end{algorithm2e}

\begin{algorithm2e}[H]
\caption{Flow Matching inference with Cremer-Pople coordinate ($\mathbf{x}_1$) prediction}\label{alg:flow_inference}
\KwIn{Prior \( p_0 \), number of integration steps \( T \), trained vector field \( u_\theta \)}
\textbf{Initialize:} \( \Delta t \leftarrow 1 / T \)\;
\( \mathbf{x}_0 \sim p_0(\mathbf{x}_0) \)\;
\( \mathbf{x}_t \leftarrow \mathbf{x}_0 \)\;
\For{$t \gets 0$  \KwTo $T-\Delta t$ \KwBy $\Delta t$} {
    \( \tilde{\mathbf{x}}_1 \leftarrow u_\theta(\mathbf{x}_t, t) \)\;
    \( \mathbf{x}_t \leftarrow \mathbf{x}_t + \Delta t (\tilde{\mathbf{x}}_1 - \mathbf{x}_t) / (1 - t) \)\
}
\Return \( \mathbf{x}_t \)

\end{algorithm2e}

\subsection{Model Architecture}\label{SI:model_architecture}

Our Equivariant Graph Neural Network follows the framework introduced by Thomas et al.~\cite{thomas_tensor_2018}. Here, $\otimes_w$ denotes the weighted Clebsch–Gordan tensor product, $\oplus$ indicates vector addition, and $\|$ concatenation. The implementation utilizes the \verb|e3nn| Python library~\cite{geiger_e3nn_2022}, similarly to \cite{jing_torsional_2022}.

\subsubsection{Embedding Layer}\label{SI:embedding_layer}
Each molecular cycle is represented as an undirected graph with atoms as nodes $\mathcal{V}$ and bonds as edges $\mathcal{E}$. Initial node embeddings $\mathbf{h}^0$ and edge embeddings $e$ are computed, where the superscript 0 indicates their scalar nature (representation of order $l=0$). 
Node features $f_a$ comprise atom-level chemical descriptors --- atomic number, hybridization, implicit valence, and ring size --- along with atom indices corresponding to Cremer--Pople numbering. These are concatenated with a sinusoidal time embedding $\phi(t)$ and processed by a learnable two-layer MLP $\Upsilon^{(v)}$ to obtain the node embeddings:
\begin{equation}
    \mathbf{h}^0_a = \Upsilon^{(v)}(f_a \, \| \, \phi(t)), \quad \forall a \in \mathcal{V}.
\end{equation}

To incorporate spatial interactions, a radius graph $\mathcal{E}_{r_{\max}}$ is constructed by adding all atom pairs within a cutoff distance $r_{\max} = 5~\text{\AA}$:
\begin{equation}
    \mathcal{E}_{r_{\max}} = \mathcal{E} \sqcup \{(a,b) \mid r_{ab} < r_{\max}\}.
\end{equation}
For each edge $(a,b) \in \mathcal{E}_{r_{\max}}$, the edge features $f_{e_{ab}}$ correspond to the bond type for the input-defined bonded pairs and are set to zero otherwise. These features are concatenated with a radial embedding $\mu(r_{ab})$, where $r_{ab}$ denotes the interatomic distance, and the same time embedding $\phi(t)$. The resulting vector is passed through a separate learnable two-layer MLP $\Upsilon^{(e)}$:
\begin{equation}
    e_{ab} = \Upsilon^{(e)}(f_{e_{ab}} \, \| \, \mu(r_{ab}) \, \| \, \phi(t)), \quad \forall (a,b) \in \mathcal{E}_{r_{\max}}.
\end{equation}

\subsubsection{Interaction Layers}\label{SI:interaction_layers}
Each interaction layer updates the node representations through batch-normalized ($\mathrm{BN}$) message passing based on Clebsch-Gordan tensor products. Specifically, messages are computed between the weighted spherical harmonic representation $Y(\hat{r}_{ab})$ of the normalized edge vector $\hat{r}_{ab}$ (up to $l = 2$) and the neighboring node embeddings $\mathbf{h}_b$, where the neighborhood of atom $a$ is defined as $\mathcal{N}_a = \{ b \mid (a,b) \in \mathcal{E}_{r_{\text{max}}} \}$. The tensor product weights $\psi_{ab}$ are predicted by a learnable MLP $\Psi$ based on the edge features $e_{ab}$ and the scalar components of the node embeddings. The node updates are thus given by:
\vspace{5px}
\begin{align}
\mathbf{h}_a & \gets \mathbf{h}_a \oplus \mathrm{BN} \Bigg( 
\frac{1}{|\mathcal{N}_a|} 
\sum_{b \in \mathcal{N}_a} 
Y(\hat{r}_{ab}) \otimes_{\psi_{ab}} \mathbf{h}_b 
\Bigg), \\
& \text{where } 
\psi_{ab} = \Psi(e_{ab}, \mathbf{h}_a^0, \mathbf{h}_b^0). \nonumber
\end{align}

\subsubsection{Cyclic Fourier Filter}\label{SI:cyclic_fourier_filter}
In the cyclic Fourier filter, atomic positions are first projected onto the mean molecular plane to obtain planar coordinates $a_m = (a_x, a_y, 0)$. For each atom $a$, an equivariant displacement embedding $\mathbf{z}_a$ is computed via message passing over its neighbors $\mathcal{N}_a$, again using Clebsch-Gordan tensor products. The embedding is defined as:
\vspace{5px}
\begin{align}
\mathbf{z}_a & \gets 
\mathrm{BN} \Bigg(
\frac{1}{|\mathcal{N}_a|} 
\sum_{b \in \mathcal{N}_a} 
Y(\hat{r}_{a_m b}) 
\otimes_{\gamma_{a_m b}} 
\mathbf{h}_b 
\Bigg), \nonumber \\
& \text{where } 
\gamma_{a_m b} = 
\Gamma\big( \mu(r_{a_m b}), \mathbf{h}_b^0 \big).
\end{align}

Here, $Y(\hat{r}_{a_m b})$ are the spherical harmonics of the projected unit vector $\hat{r}_{a_m b}$, and $r_{a_m b}$ is the corresponding distance between atoms $a$ and $b$ in the mean plane. The tensor product weights $\gamma_{a_m b}$ are produced by the MLP $\Gamma$, conditioned on the radial basis expansion $\mu(r_{a_m b})$ and the scalar node features $\mathbf{h}_b^0$.


The scalar components of the resulting embedding $\mathbf{z}_a$ are subsequently processed by a bias-free MLP with $\tanh$ activations to preserve equivariance, yielding the final atomic displacements $z_a$. These displacements are then transformed via a scaled discrete Fourier transform to obtain the predicted Cremer-Pople coordinates.

\subsubsection{Generating Exocyclic Substituents}\label{SI:exocyclic}
While \name{} generates cyclic substructures, it allows for complete molecular conformer generation when combined with established methods that model exocyclic substituents, such as RDKit's ETKDG \cite{landrum_rdkitrdkit_2024} or torsional diffusion \cite{jing_torsional_2022}. 
For integration with RDKit, we keep the generated rings as a fixed substructure and generate the exocyclic substituents as usual. For integration with torsional diffusion, we sample the ring structures using \name{}, and use these as local substructures for a pretrained model of torsional diffusion using publicly available model weights \cite{jing_torsional_2022}. Future work aims at generating both torsional and cyclic degrees of freedom simultaneously, which is feasible due to the compatibility of the architectures of \name{} and torsional diffusion.

\subsection{Benchmarking Methodology}\label{SI:benchmarking_methodology}

We follow the evaluation metrics for conformer generation introduced by Ganea \textit{et al.} \cite{ganea_geomol_2021} and following works. For each molecule, we calculate the Average Minimum RMSD (AMR) and Coverage (COV) for Precision (P) and Recall (R) with the following formulas

\begin{align}
\text{AMR-R} &:=  \frac{1}{M} \sum_{m \in [1..M]} \Bigg[\frac{1}{L} \sum_{l \in [1..L_m]} \min_{k \in [1..K_m]} \text{RMSD}(C_{k,m}, C_{l,m}^*)\Bigg] \\
\text{AMR-P} &:= \frac{1}{M} \sum_{m \in [1..M]} \Bigg[\frac{1}{K} \sum_{k \in [1..K_m]} \min_{l \in [1..L_m]} \text{RMSD}(C_{k,m}, C_{l,m}^*)\Bigg] \\
\text{COV-R} &:= \frac{1}{M} \sum_{m \in [1..M]} \Bigg[ \frac{1}{L} |\{l \in [1..L_m]: \exists k \in [1..K_m],\ \text{RMSD}(C_{k,m}, C_{l,m}^*) < \delta\}|\Bigg]\\
\text{COV-P} &:= \frac{1}{M} \sum_{m \in [1..M]} \Bigg[ \frac{1}{K} |\{k \in [1..K_m]: \exists l \in [1..L_m],\ \text{RMSD}(C_{k,m}, C_{l,m}^*) < \delta\}|\Bigg]
\end{align}

where $l$ denotes a conformer in the dataset ($\{C_{l,m}^*\}_{l\in[1,L_m]}$) and $k$ a conformer generated with our model ($\{C_{k,m}\}_{k\in[1,K_m]}$) for molecule $m$, with $M$ the total number of molecules. For each molecule with $L$ conformers in the test set, $K=\text{min}(50,2L)$ conformers are generated. For MCF, we benchmark against the small model version (MCF-S), as the number of parameters (13M) is closest to the one of \name{} (966K), while still being more than a magnitude larger.

\subsection{Additional Results}\label{SI:additional_results}

\subsubsection{Hyperparameter Optimization}\label{SI:hyperparameter_optimization}
We trained on an Nvidia GeForce RTX 4090 GPU for 300 epochs with the AdamW optimizer (taking approx. 1 hour). The hyperparameters tuned on the validation set were: initial learning rate (0.0005, \textbf{0.001}, 0.005), number of layers (2, \textbf{4}, 6), maximum representation order (1st, \textbf{2nd}), batch norm (\textbf{True}, False), and the sizes of the internal representations for first (\textbf{32}, 16) and second order (\textbf{4}, 2). 

\subsubsection{Standard errors between data splits}\label{SI:sems}

Standard errors of the mean, calculated across five independent data splits, are reported in the tables below for the same metrics reported in the main text. Ring puckering results are shown in Tab.~\ref{tab:results_rmse_random_split_sems}, followed by atomic position RMSD results in Tab.~\ref{tab:results_rmsd_random_split_sems}.

\subsubsection{Effect of Inference Step Count}\label{SI:inference_step_count}
We assessed the effect of the number of inference steps on model performance (see Tabs.~\ref{tab:rmse_inference_steps}--\ref{tab:rmsd_inference_steps}). For each step count, the mean values averaged over the five different data splits are shown. The results indicate that the model achieves comparable performance when performing only 2-5 inference steps versus the 30 steps presented for \name{} in the main text.

\begin{landscape}

\begin{table}
  \centering
  \small 
  \begin{tabular}{lcccccccc}
    \multirow{3}{*} & \multicolumn{4}{c}{(a) Unrelaxed} & \multicolumn{4}{c}{(b) Relaxed with MMFF} \\
    \cmidrule(lr){2-5} \cmidrule(lr){6-9}
    & \multicolumn{2}{c}{Precision} & \multicolumn{2}{c}{Recall} & \multicolumn{2}{c}{Precision} & \multicolumn{2}{c}{Recall} \\
    \cmidrule(lr){2-3} \cmidrule(lr){4-5} \cmidrule(lr){6-7} \cmidrule(lr){8-9}
    Method & \shortstack{AMR\\(\AA) $\downarrow$} & \shortstack{Coverage\\(\%) $\uparrow$} & \shortstack{AMR \\ (\AA)$ \downarrow$} & \shortstack{Coverage\\(\%) $\uparrow$} & \shortstack{AMR\\(\AA) $\downarrow$} & \shortstack{Coverage\\(\%) $\uparrow$} & \shortstack{AMR \\ (\AA) $\downarrow$} & \shortstack{Coverage\\(\%) $\uparrow$} \\
    \cmidrule(l){1-1} \cmidrule(lr){2-5} \cmidrule(lr){6-9}
    \textbf{\name{} (ours)}    & 0.13 $\pm$ 0.008 & 67.5 $\pm$ 1.2 & 0.09 $\pm$ 0.008 & 75.8 $\pm$ 0.6 & 0.13 $\pm$ 0.009 & 70.8 $\pm$ 3.2 & 0.11 $\pm$ 0.009 & 72.2 $\pm$ 3.9 \\
RDKit ETKDG (Small Cycles) & 0.17 $\pm$ 0.011 & 51.4 $\pm$ 2.0 & 0.13 $\pm$ 0.008 & 60.1 $\pm$ 1.8 & 0.15 $\pm$ 0.010 & 63.4 $\pm$ 2.4 & 0.12 $\pm$ 0.009 & 65.5 $\pm$ 3.7 \\
RDKit ETKDG & 0.18 $\pm$ 0.011 & 43.6 $\pm$ 3.6 & 0.14 $\pm$ 0.010 & 53.3 $\pm$ 2.7 & 0.15 $\pm$ 0.012 & 61.0 $\pm$ 2.6 & 0.13 $\pm$ 0.012 & 63.7 $\pm$ 3.5 \\
RDKit KDG & 0.20 $\pm$ 0.011 & 33.0 $\pm$ 1.8 & 0.15 $\pm$ 0.010 & 51.4 $\pm$ 1.2 & 0.15 $\pm$ 0.012 & 60.3 $\pm$ 1.9 & 0.13 $\pm$ 0.010 & 64.0 $\pm$ 2.8 \\
MCF & 0.16 $\pm$ 0.011 & 46.2 $\pm$ 1.6 & 0.12 $\pm$ 0.007 & 60.0 $\pm$ 1.2 & 0.14 $\pm$ 0.010 & 64.3 $\pm$ 1.8 & 0.11 $\pm$ 0.007 & 68.1 $\pm$ 3.3 \\
GeoDiff & 0.24 $\pm$ 0.016 & 24.9 $\pm$ 1.5 & 0.18 $\pm$ 0.018 & 42.4 $\pm$ 3.2 & 0.17 $\pm$ 0.016 & 57.2 $\pm$ 2.3 & 0.14 $\pm$ 0.016 & 61.4 $\pm$ 3.5 \\
    \cmidrule(l){1-1} \cmidrule(lr){2-5} \cmidrule(lr){6-9}
  \end{tabular}
  \vspace{2px}
  \caption{\centering{Quantitative benchmarking on the ring puckering,  \textit{i.e.}, the atomic displacements from the mean plane (see Fig.~\ref{fig:methods_model_overview}a), including standard errors. We report the RMSD precision and recall for (a) unrelaxed and (b) relaxed structures in terms of Average Minimum RMSD (AMR) and coverage (0.1~\AA{} threshold).}}
  \label{tab:results_rmse_random_split_sems}
\end{table}

\end{landscape}
\begin{landscape}

\begin{table}
  \centering
  \small 
  \begin{tabular}{lcccccccc}
    \multirow{3}{*} & \multicolumn{4}{c}{(a) Unrelaxed} & \multicolumn{4}{c}{(b) Relaxed with MMFF} \\
    \cmidrule(lr){2-5} \cmidrule(lr){6-9}
    & \multicolumn{2}{c}{Precision} & \multicolumn{2}{c}{Recall} & \multicolumn{2}{c}{Precision} & \multicolumn{2}{c}{Recall} \\
    \cmidrule(lr){2-3} \cmidrule(lr){4-5} \cmidrule(lr){6-7} \cmidrule(lr){8-9}
    Method & \shortstack{AMR\\(\AA) $\downarrow$} & \shortstack{Coverage\\(\%) $\uparrow$} & \shortstack{AMR \\ (\AA)$ \downarrow$} & \shortstack{Coverage\\(\%) $\uparrow$} & \shortstack{AMR\\(\AA) $\downarrow$} & \shortstack{Coverage\\(\%) $\uparrow$} & \shortstack{AMR \\ (\AA) $\downarrow$} & \shortstack{Coverage\\(\%) $\uparrow$} \\
    \cmidrule(l){1-1} \cmidrule(lr){2-5} \cmidrule(lr){6-9}
    \textbf{\name{} (ours)} & 0.18 $\pm$ 0.009 & 47.4 $\pm$ 2.1 & 0.15 $\pm$ 0.010 & 51.0 $\pm$ 1.4 & 0.16 $\pm$ 0.009 & 61.5 $\pm$ 2.0 & 0.14 $\pm$ 0.010 & 61.0 $\pm$ 2.8 \\
RDKit ETKDG (Small Cycles) & 0.22 $\pm$ 0.011 & 37.7 $\pm$ 2.9 & 0.17 $\pm$ 0.009 & 44.5 $\pm$ 2.8 & 0.17 $\pm$ 0.009 & 55.8 $\pm$ 0.8 & 0.14 $\pm$ 0.009 & 56.3 $\pm$ 2.6 \\
RDKit ETKDG & 0.22 $\pm$ 0.012 & 35.6 $\pm$ 3.5 & 0.17 $\pm$ 0.010 & 42.9 $\pm$ 3.3 & 0.18 $\pm$ 0.011 & 54.2 $\pm$ 1.8 & 0.15 $\pm$ 0.010 & 54.6 $\pm$ 3.0 \\
RDKit KDG & 0.23 $\pm$ 0.012 & 26.4 $\pm$ 2.4 & 0.18 $\pm$ 0.010 & 41.5 $\pm$ 2.1 & 0.18 $\pm$ 0.011 & 54.3 $\pm$ 0.7 & 0.16 $\pm$ 0.009 & 55.7 $\pm$ 2.0 \\
MCF & 0.23 $\pm$ 0.015 & 24.6 $\pm$ 1.7 & 0.19 $\pm$ 0.013 & 35.6 $\pm$ 3.1 & 0.17 $\pm$ 0.009 & 56.4 $\pm$ 0.9 & 0.14 $\pm$ 0.007 & 58.2 $\pm$ 2.2 \\
GeoDiff & 0.28 $\pm$ 0.017 & 16.4 $\pm$ 1.3 & 0.22 $\pm$ 0.018 & 30.5 $\pm$ 2.3 & 0.20 $\pm$ 0.016 & 50.9 $\pm$ 1.6 & 0.17 $\pm$ 0.017 & 53.9 $\pm$ 2.6 \\
    \cmidrule(l){1-1} \cmidrule(lr){2-5} \cmidrule(lr){6-9}
  \end{tabular}
  \vspace{2px}
  \caption{\centering{Quantitative benchmarking on the atomic position RMSD including standard errors. We report precision and recall for (a) unrelaxed and (b) relaxed structures in terms of Average Minimum RMSD (AMR) and coverage (0.1~\AA{} threshold).}}
  \label{tab:results_rmsd_random_split_sems}
\end{table}             

\end{landscape}

\begin{landscape}

\begin{table}
  \centering
  \renewcommand{\arraystretch}{1.5}
  \resizebox{1.6\textwidth}{!}{%
  \begin{tabular}{lcccccccccccccccccccccccccccc}
    \toprule
    & \multicolumn{14}{c}{Precision} & \multicolumn{14}{c}{Recall} \\
    \cmidrule(lr){2-15} \cmidrule(lr){16-29}
    & \multicolumn{7}{c}{Average Mean RMSD (\AA{}) $\downarrow$} & \multicolumn{7}{c}{Coverage (\%) $\uparrow$} & \multicolumn{7}{c}{Average Mean RMSD (\AA{}) $\downarrow$} & \multicolumn{7}{c}{Coverage (\%) $\uparrow$} \\
    \cmidrule(lr){2-8} \cmidrule(lr){9-15} \cmidrule(lr){16-22} \cmidrule(lr){23-29}
    Method & 1 & 2 & 5 & 10 & 20 & 30 & 50 & 1 & 2 & 5 & 10 & 20 & 30 & 50 & 1 & 2 & 5 & 10 & 20 & 30 & 50 & 1 & 2 & 5 & 10 & 20 & 30 & 50 \\
    \midrule
    \name{} (\textit{unrel}) & \gradientAMR{0.16} & \gradientAMR{0.13} & \gradientAMR{0.13} & \gradientAMR{0.13} & \gradientAMR{0.13} & \gradientAMR{0.13} & \gradientAMR{0.13} & \gradientCov{36.6} & \gradientCov{54.8} & \gradientCov{64.9} & \gradientCov{67.2} & \gradientCov{67.4} & \gradientCov{67.5} & \gradientCov{67.5} & \gradientAMR{0.18} & \gradientAMR{0.11} & \gradientAMR{0.09} & \gradientAMR{0.09} & \gradientAMR{0.09} & \gradientAMR{0.09} & \gradientAMR{0.09} & \gradientCov{29.4} & \gradientCov{63.1} & \gradientCov{74.4} & \gradientCov{76.2} & \gradientCov{76.2} & \gradientCov{75.8} & \gradientCov{74.9} \\
\name{} (\textit{rel}) & \gradientAMR{0.13} & \gradientAMR{0.13} & \gradientAMR{0.13} & \gradientAMR{0.13} & \gradientAMR{0.13} & \gradientAMR{0.13} & \gradientAMR{0.13} & \gradientCov{67.9} & \gradientCov{70.1} & \gradientCov{70.4} & \gradientCov{70.5} & \gradientCov{70.6} & \gradientCov{70.8} & \gradientCov{70.5} & \gradientAMR{0.12} & \gradientAMR{0.11} & \gradientAMR{0.11} & \gradientAMR{0.10} & \gradientAMR{0.11} & \gradientAMR{0.11} & \gradientAMR{0.10} & \gradientCov{64.8} & \gradientCov{71.1} & \gradientCov{71.8} & \gradientCov{71.8} & \gradientCov{72.2} & \gradientCov{72.2} & \gradientCov{71.6} \\
\bottomrule
    \bottomrule
  \end{tabular}%
  }
  \vspace{5px}
  \caption{Quantitative benchmarking across inference steps on the ring puckering, \textit{i.e.}, the atomic displacements from the mean plane. We report precision and recall for both unrelaxed and relaxed structures in terms of Average Mean RMSD (AMR) and coverage (0.1 \AA{} threshold). Colors represent normalized AMR (per category, lighter green indicates better performance) and coverage (green = best, red = worst).}
  \label{tab:rmse_inference_steps}
\end{table}

\begin{table}
  \centering
  \renewcommand{\arraystretch}{1.5}
  \resizebox{1.6\textwidth}{!}{%
  \begin{tabular}{lcccccccccccccccccccccccccccc}
    \toprule
    & \multicolumn{14}{c}{Precision} & \multicolumn{14}{c}{Recall} \\
    \cmidrule(lr){2-15} \cmidrule(lr){16-29}
    & \multicolumn{7}{c}{Average Mean RMSD (\AA{}) $\downarrow$} & \multicolumn{7}{c}{Coverage (\%) $\uparrow$} & \multicolumn{7}{c}{Average Mean RMSD (\AA{}) $\downarrow$} & \multicolumn{7}{c}{Coverage (\%) $\uparrow$} \\
    \cmidrule(lr){2-8} \cmidrule(lr){9-15} \cmidrule(lr){16-22} \cmidrule(lr){23-29}
    Method & 1 & 2 & 5 & 10 & 20 & 30 & 50 & 1 & 2 & 5 & 10 & 20 & 30 & 50 & 1 & 2 & 5 & 10 & 20 & 30 & 50 & 1 & 2 & 5 & 10 & 20 & 30 & 50 \\
    \midrule
    \name{} (\textit{unrel}) & \gradientAMR{0.22} & \gradientAMR{0.19} & \gradientAMR{0.18} & \gradientAMR{0.18} & \gradientAMR{0.18} & \gradientAMR{0.18} & \gradientAMR{0.18} & \gradientCov{24.3} & \gradientCov{35.1} & \gradientCov{43.8} & \gradientCov{45.9} & \gradientCov{47.2} & \gradientCov{47.4} & \gradientCov{47.3} & \gradientAMR{0.23} & \gradientAMR{0.16} & \gradientAMR{0.14} & \gradientAMR{0.14} & \gradientAMR{0.15} & \gradientAMR{0.15} & \gradientAMR{0.15} & \gradientCov{20.1} & \gradientCov{39.6} & \gradientCov{48.8} & \gradientCov{50.5} & \gradientCov{51.0} & \gradientCov{51.0} & \gradientCov{49.9} \\
\name{} (\textit{rel}) & \gradientAMR{0.16} & \gradientAMR{0.16} & \gradientAMR{0.16} & \gradientAMR{0.16} & \gradientAMR{0.16} & \gradientAMR{0.16} & \gradientAMR{0.16} & \gradientCov{59.5} & \gradientCov{61.4} & \gradientCov{61.4} & \gradientCov{61.5} & \gradientCov{61.5} & \gradientCov{61.5} & \gradientCov{61.6} & \gradientAMR{0.15} & \gradientAMR{0.14} & \gradientAMR{0.14} & \gradientAMR{0.14} & \gradientAMR{0.14} & \gradientAMR{0.14} & \gradientAMR{0.14} & \gradientCov{56.2} & \gradientCov{60.5} & \gradientCov{60.8} & \gradientCov{61.0} & \gradientCov{61.0} & \gradientCov{61.0} & \gradientCov{60.8} \\
    \bottomrule
  \end{tabular}}
  \vspace{5px}
  \caption{Quantitative benchmarking across inference steps on atomic position RMSD. We report precision and recall for both unrelaxed and relaxed structures in terms of Average Mean RMSD (AMR) and coverage (0.1 \AA{} threshold). Colors represent normalized AMR (per category, lighter green indicates better performance) and coverage (green = best, red = worst).}
  \label{tab:rmsd_inference_steps}
\end{table}
\end{landscape}

\newpage
\printbibliography

\end{refsection}

\end{document}